\documentclass[manuscript,nonacm]{acmart}
\usepackage{algorithm}
\usepackage{algorithmic}
\usepackage{amsmath}
\usepackage{soul}
\usepackage[utf8]{inputenc}
\usepackage{xcolor}
\usepackage{booktabs}
\usepackage{multirow}
\usepackage{xparse,mleftright}
\usepackage{marvosym}
\usepackage{CJKutf8}
\usepackage{makecell}

\usepackage{color}
\definecolor{BurntOrange}{rgb}{0.8, 0.33, 0}
\definecolor{OliveGreen}{RGB}{85, 107, 47}
\definecolor{OliveGreen}{RGB}{85, 107, 47}
\definecolor{Plum}{RGB}{139, 102, 139}

\AtBeginDocument{%
  }

\setcopyright{acmlicensed}
\copyrightyear{2025}
\acmYear{2025}
\acmDOI{XXXXXXX.XXXXXXX}

\acmJournal{JACM}
\acmVolume{1}
\acmNumber{1}
\acmArticle{1}
\acmMonth{6}

\begin{document}

\title{Ancient Script Image Recognition and Processing: A Review}

\author{Xiaolei Diao}
\affiliation{%
  \institution{Queen Mary University of London}
  \city{London}
  \country{UK}}
\email{x.diao@qmul.ac.uk}

\author{Rite Bo}
\affiliation{%
  \institution{Jilin University}
  \city{Changchun}
  \country{China}}
\email{bort24@mails.jlu.edu.cn}

\author{Yanling Xiao}
\affiliation{%
  \institution{Jilin University}
  \city{Changchun}
  \country{China}}
\email{xiaoyl24@mails.jlu.edu.cn}

\author{Lida Shi}
\affiliation{%
  \institution{Jilin University}
  \city{Changchun}
  \country{China}}
\email{shild21@mails.jlu.edu.cn}

\author{Zhihan Zhou}
\affiliation{%
  \institution{Jilin University}
  \city{Changchun}
  \country{China}}
\email{zhzhou22@mails.jlu.edu.cn}

\author{Hao Xu}
\affiliation{%
  \institution{Jilin University}
  \city{Changchun}
  \country{China}}
\email{xuhao@jlu.edu.cn}

\author{Chuntao Li}
\affiliation{%
  \institution{Jilin University}
  \city{Changchun}
  \country{China}}
\email{lct33@jlu.edu.cn}

\author{Xiongfeng Tang}
\affiliation{%
  \institution{Jilin University}
  \city{Changchun}
  \country{China}}
\email{tangxf921@jlu.edu.cn}

\author{Massimo Poesio}
\affiliation{%
  \institution{Queen Mary University of London}
  \city{London}
  \country{UK}}
\email{m.poesio@qmul.ac.uk}

\author{Cédric M. John}
\affiliation{%
  \institution{Queen Mary University of London}
  \city{London}
  \country{UK}}
\email{cedric.john@qmul.ac.uk}

\author{Daqian Shi}
\affiliation{%
  \institution{Queen Mary University of London}
  \city{London}
  \country{UK}}
\email{d.shi@qmul.ac.uk}
\authornote{Corresponding author.}

\renewcommand{\shortauthors}{Diao et al.}

\begin{abstract}

Ancient scripts, e.g., Egyptian hieroglyphs, Oracle Bone Inscriptions, and Ancient Greek inscriptions, serve as vital carriers of human civilization, embedding invaluable historical and cultural information. Automating ancient script image recognition has gained importance, enabling large-scale interpretation and advancing research in archaeology and digital humanities. With the rise of deep learning, this field has progressed rapidly, with numerous script-specific datasets and models proposed. While these scripts vary widely, spanning phonographic systems with limited glyphs to logographic systems with thousands of complex symbols, they share common challenges and methodological overlaps. Moreover, ancient scripts face unique challenges, including imbalanced data distribution and image degradation, which have driven the development of various dedicated methods. This survey provides a comprehensive review of ancient script image recognition methods. We begin by categorizing existing studies based on script types and analyzing respective recognition methods, highlighting both their differences and shared strategies. We then focus on challenges unique to ancient scripts, systematically examining their impact and reviewing recent solutions, including few-shot learning and noise-robust techniques. Finally, we summarize current limitations and outline promising future directions. Our goal is to offer a structured, forward-looking perspective to support ongoing advancements in the recognition, interpretation, and decipherment of ancient scripts.

\end{abstract}




% \received{23 June 2025}

\maketitle

\section{Introduction}
\label{sec:Introduction}
Fragile artifacts and faded inscriptions house the understanding of past civilizations, with notorious examples including the Rosetta Stone and the Dead Sea Scrolls. The key to this knowledge is ancient scripts, such as ancient Greek inscriptions, Egyptian hieroglyphs, and Oracle Bone Inscriptions (OBIs). These archaic writing systems bear witness to pivotal moments in human history and trace the lineage of civilization \cite{luo2021deciphering}. By deciphering ancient scripts, scholars gain profound insights into the political, economic, and religious dimensions of past societies, highlighting civilization roots with impact into the modern world \cite{assael2022restoring}. Consequently, researching ancient scripts is of great importance not only for historical inquiry but also for the preservation of cultural heritage. As principal digital resources, ancient script images are key research objects for scientists\footnote{To preserve fragile ancient artifacts and facilitate purposes such as readability and data processing, ancient scripts are often transformed into images via digitization methods.}. Automated recognition and processing of these digitized images promises to enable the large-scale interpretation of historical scripts \cite{ferrara2022advanced}.

\begin{figure}[!t]
  \centering
  \includegraphics[width=0.95\linewidth]{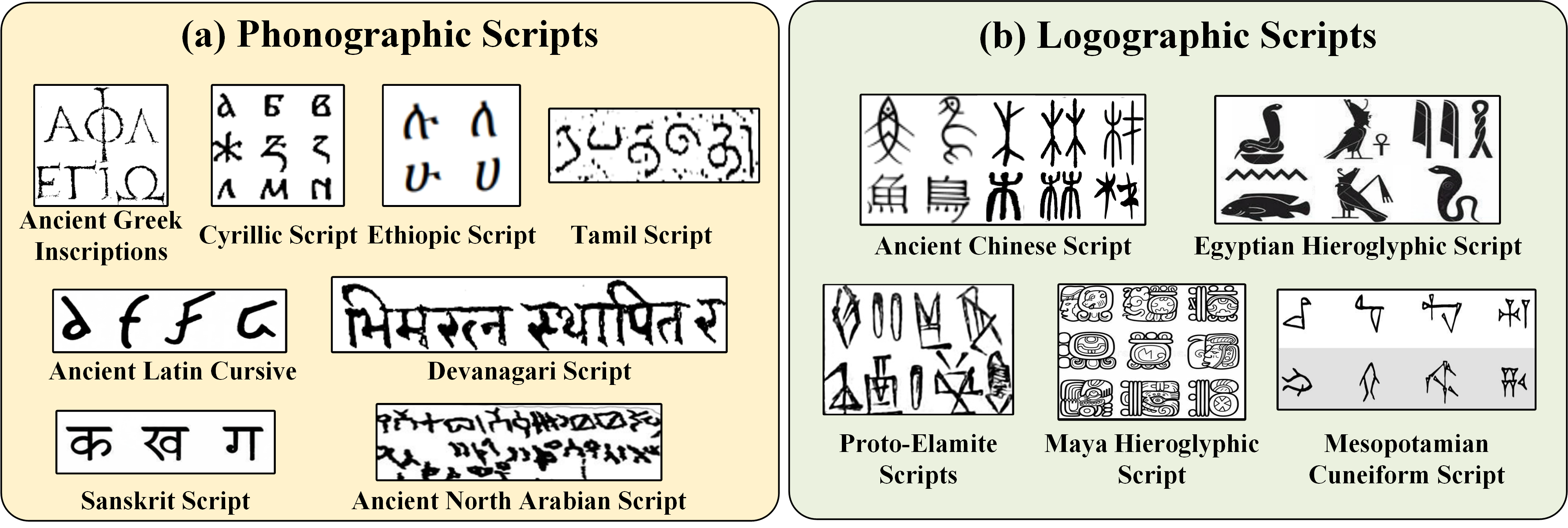}
  \caption{Examples of different ancient scripts as phonographic scripts and logographic scripts.}
    \label{img1}
\end{figure}

But ancient scripts are diverse with different characteristics and challenges \cite{ischebeck2004processing}. Based on the definition by linguists, widely studied ancient scripts can be categorized into (1) Phonographic scripts, such as Latin, Ancient Greek, and Sanskrit; and (2) Logographic scripts, such as OBIs, Sumerian Cuneiform, and Egyptian Hieroglyphs \cite{liu1995script}. These two types of scripts represent distinct approaches to linguistic encoding, resulting in shared data characteristics among scripts within the same type \cite{krithiga2023ancient}. Fig.~\ref{img1} illustrates examples of images from both types of ancient scripts. Phonographic scripts encode segmental pronunciations into characters, each closely linked to the phonological structure of the language. Phonographic scripts generally contain a relatively small number of characters, characterized by simpler glyphs \cite{gnanadesikan2017towards}. Thus, characters need to be formed as words to express different meanings. On the other hand, logographic scripts focus on conveying semantics through individual characters, often employing symbols to represent objects or concepts \cite{gnanadesikan2017towards}. This results in a larger set of unique characters in logographic scripts, with intricate glyphs that represent complex meanings.

\begin{figure}[!t]
  \centering
\includegraphics[width=0.5\linewidth]{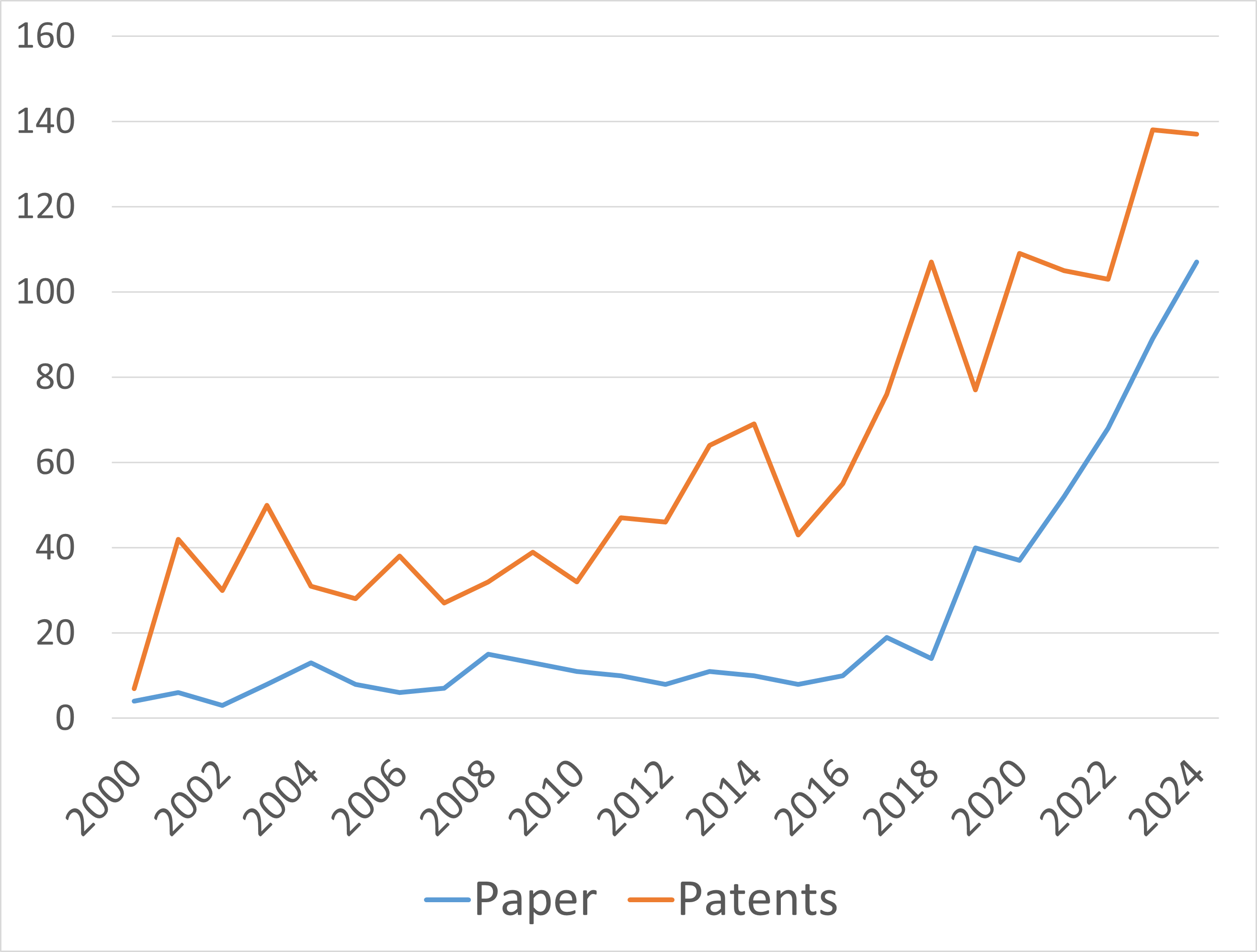}
  \caption{Annual cumulative counts of published papers and granted patents for ancient script recognition using deep learning from 2000 to 2024.}
    \label{img1.1}
\end{figure}

Recognizing ancient script images relies on manual analysis or computer vision-based image processing tools. Experts manually analyze, classify, and interpret each character, making this process not only time-consuming but also prone to errors \cite{shen2023challenges}. Recent advances in artificial intelligence (AI), particularly optical character recognition (OCR) algorithms based on deep learning (DL), offer new opportunities to address these challenges \cite{hirugade2022survey}. In recent years, numerous DL- and computer vision-based methods for ancient script recognition have emerged, aiming to perform classification, recognition, and interpretation tasks at a larger scale and with higher accuracy \cite{al2020handwritten}. Fig.~\ref{img1.1} shows the number of publications and patents related to ancient script recognition from 2000 to 2024\footnote{The reference data are sourced from Google Scholar, Scopus, and The Lens.}, revealing a pronounced upward trend, with a substantial number of scholarly papers published in the last five years. Given that many commonalities are shared in the recognition of different ancient scripts, we are motivated to undertake a more systematic review and synthesis of these studies, rather than studying each script in isolation \cite{yokoi2024phonology}.

Some DL-based methods for recognizing ancient scripts have shown the feasibility of this approach; however, they also face challenges when applied to real-world scenarios. These challenges mainly arise from two aspects that distinguish ancient scripts from common modern scripts: (1) artifacts unearthed after thousands of years are often degraded or damaged, resulting in images containing significant real-world noise \cite{shi2022charformer}; and (2) the scarcity of unearthed artifacts results in an insufficient number of ancient script samples \cite{diao2023toward}. These challenges degrade the quality of ancient script image datasets, making it difficult to meet the training requirements of DL-based models and ultimately leading to suboptimal recognition performance \cite{narang2020ancient}. Consequently, many studies have attempted various methods to overcome these problems. Currently, there remains a lack of reviews that specifically address the unique challenges associated with ancient script images.

In this paper, our objective is to conduct a systematic review of ancient script image recognition studies from multiple perspectives. Based on existing studies, we provide an in-depth discussion of the two major types of ancient scripts (logographic and phonographic), highlighting both the distinctions and interrelations between their respective image recognition approaches. Additionally, we provide an innovative review of two unique challenges specific to ancient script images: the presence of natural noise and the scarcity of available samples, discussing the resulting issues as well as corresponding solutions. This study also examines the limitations of current methodologies and explores potential directions for future research, offering researchers a broader perspective through meaningful observations. Our contributions can be summarized as follows:

\begin{itemize}
    \item We incorporate the latest methods for image recognition, providing an overview of recent advancements in ancient script recognition. We systematically present recognition approaches for 17 ancient scripts, primarily categorized as logographic and phonographic, discussing their interrelated research from a linguistic perspective. 

    \item We conduct an in-depth examination of the application of few-shot learning methods in ancient script recognition, exploring strategies employed by related methods to effectively learn from very few samples using few-shot or zero-shot algorithms. We also introduce the issue of noise in ancient script image data, discussing the impact of different types of noise on recognition performance and presenting methods for accurate recognition in noisy images.

    \item We provide a visual summary in tabular form to compare our survey with previous ones. We highlight the areas addressed by previous studies and those that have not yet been thoroughly discussed, establishing the unique contributions of this review. Additionally, we discuss the limitations of current methods and potential areas for future research.
\end{itemize}

The remainder of this survey is structured as follows: Section 2 introduces the types and characteristics of ancient scripts under consideration, clarifying how our review differs from existing surveys. Section 3 discusses methods tailored to recognizing logographic scripts, followed by Section 4, which deals with phonographic script recognition. Section 5 examines the problem of few-shot learning, and Section 6 focuses on the presence of noise in ancient script images. Finally, Section 7 concludes the paper and proposes potential directions for future exploration.

\section{
Revisiting Ancient Script Recognition
% Key Differentiators and Insights
% Core Motivation and Insights
}

\label{sec:Categorization}
The taxonomy of ancient scripts is crucial for organizing and learning various recognition methods, as scripts within the same category often have similar writing features. In this section, we aim to systematically discuss the data features associated with these different script categories. We will explore the unique challenges that images of these ancient scripts pose to current recognition methods. Furthermore, we will outline the core motivations driving our survey and demonstrate how it distinguishes itself from existing surveys. By providing new perspectives and addressing critical gaps in this research domain, we aim to highlight the novel contributions of our work. 

\subsection{Phonographic Scripts vs. Logographic Scripts}

Based on linguistic principles and the visual features of script writing, widely studied ancient scripts can be broadly categorized into phonographic scripts and logographic scripts \cite{ischebeck2004processing, liu1995script}. Phonographic scripts are characterized by their representation of phonemes, the smallest units of sound in a language. These scripts typically employ smaller, simpler, and relatively uniform characters designed to systematically map sounds, as illustrated in Fig.~\ref{img1}(a). Representative examples include Latin cursive scripts, where characters represent individual sounds, and abugidas such as Sanskrit, where consonant-vowel combinations are encoded within single characters \cite{ghosh2010script}. This results in visual similarities across different phonographic scripts due to their shared focus on sound representation \cite{toyoda2009common}. Specifically, \textit{phonographic scripts} exhibit the following common features:
\begin{itemize}
    \item \textbf{Small and similar characters}: The characters in phonographic scripts, especially in handwritten forms, often bear visual similarity. For instance, letters like ``e" and ``f" in Latin cursive scripts are prone to confusion due to their shapes and minimal distinguishing features. This similarity increases recognition complexity when scripts are degraded or noisy.

    \item \textbf{The recognition of words and characters}: In phonographic writing systems, successfully recognizing individual characters is not necessarily considered the completion of the recognition task. This is because the semantics in these scripts are not carried by individual characters but rather by the combination of characters that form words. Consequently, corresponding recognition methods often also consider word-level recognition to achieve higher semantic accuracy.

    \item \textbf{Hand-writing styles}: Cursive or joined-up writing styles are frequently observed in phonographic scripts. Such connected strokes introduce noise and structural variations, making character segmentation and recognition particularly challenging. This issue is even more pronounced in ancient handwritten manuscripts, where stylistic variations and quality degradation are common.
\end{itemize}

Logographic scripts fundamentally differ from phonographic scripts in that their characters represent morphemes, the smallest units of meaning \cite{daniels1996world}. These scripts are visually and structurally more complex, with characters often resembling intricate patterns rather than simple shapes \cite{ellis2004effects}. This distinction results in unique visual features for logographic scripts, setting them apart from phonographic scripts. Representative logographic scripts like OBIs and ancient Egyptian hieroglyphs can be found in Fig.~\ref{img1}(b), where each character typically conveys a specific meaning or concept, representing morphemes rather than sounds. In particular, the following characteristics are observed in \textit{logographic scripts}:
\begin{itemize}
    \item \textbf{Complex and Diverse Structures}: Individual logographic characters are typically intricate, consisting of multiple strokes or components. This structural complexity makes them challenging to recognize, particularly in degraded or incomplete images.
    
    \item \textbf{Sparse Character Set}: Since each individual logographic character corresponds to a specific semantic meaning, the number of characters in such scripts is often significantly larger than in phonographic scripts. As a result, accurately recognizing each character presents a considerable challenge.
    
    \item \textbf{Glyph Similarity}: Logographic characters often contain a large number of reusable components, known as radicals or subcomponents, which carry specific semantic meanings and are frequently reused across different characters. This results in high visual similarity among many logographic characters. For instance, in Chinese, characters such as ``\begin{CJK}{UTF8}{gbsn}材\end{CJK}" and ``\begin{CJK}{UTF8}{gbsn}林\end{CJK}" share the common component ``\begin{CJK}{UTF8}{gbsn}木\end{CJK}", making it challenging to distinguish between these characters visually\footnote{These three characters mean timber, forest, and wood, respectively.}.
\end{itemize}

\begin{figure}[!t]
  \centering
  \includegraphics[width=0.8\linewidth]{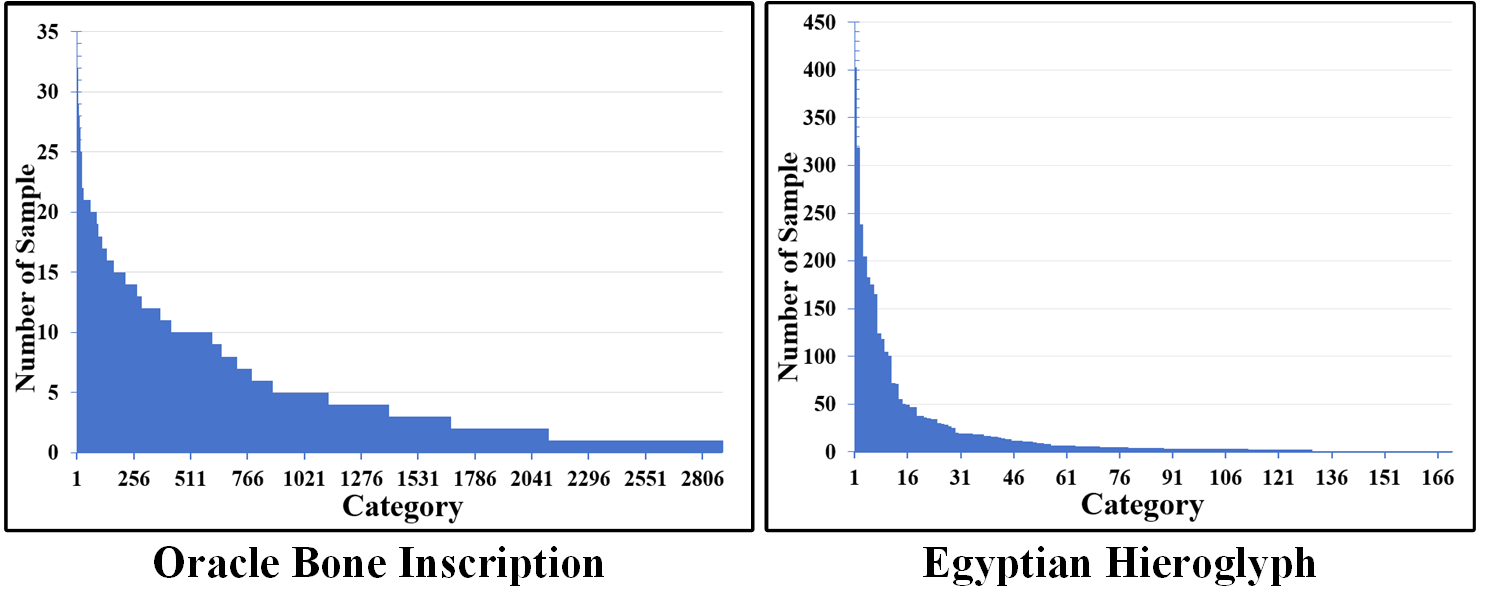}
  \caption{Representative example of the imbalanced distribution in ancient script datasets.}
    \label{img2}
\end{figure}

\subsection{Challenges of Ancient Scripts Images}

Ancient script images exhibit unique characteristics that pose significant challenges to conventional recognition methods. These challenges stem from two fundamental issues: \textit{limited sample availability} and the \textit{real world noise} in the images, both of which are deeply rooted in the historical and environmental conditions under which ancient scripts were created and preserved.

\begin{figure}[!t]
  \centering
  \includegraphics[width=0.85\linewidth]{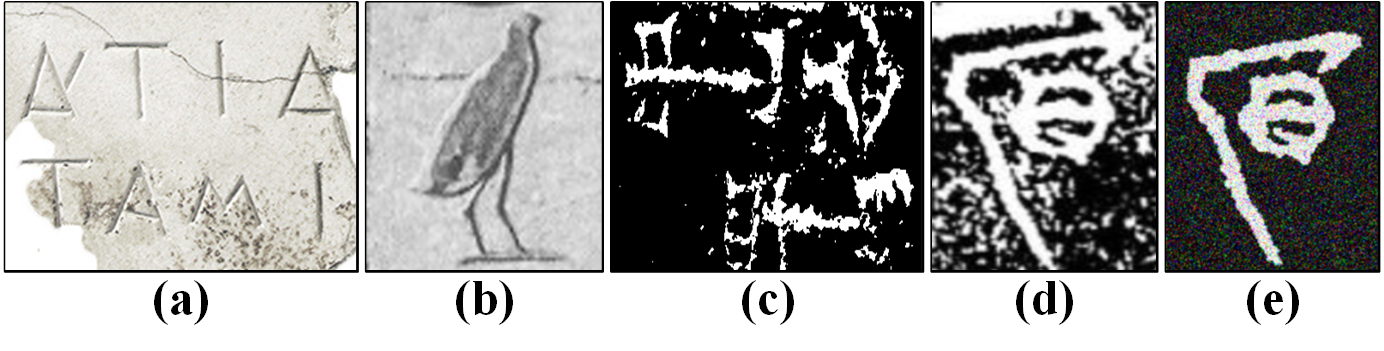}
  \caption{Representative examples of the different visual degradation on ancient scripts. (a) Greek inscriptions with cracks and damaged edges, (b) Egyptian hieroglyphs with scratches, (c) Sumerian cuneiform with erosion, (d)-(e) OBIs with erosion and synthetic noises, respectively.}
    \label{img3}
\end{figure}

One of the main challenges in ancient script recognition is the scarcity and imbalance of available data \cite{li2023towards}. Wars, natural disasters, and material degradation have significantly reduced the number of surviving manuscripts, inscriptions, and artifacts \cite{lin2025handwriting}, limiting the total number of collectible samples. Moreover, character frequencies within these datasets are often highly uneven \cite{wu2021ancient}, with some characters appearing frequently while others occur only once or twice. As shown in Fig.~\ref{img2}, both the OBI and Egyptian hieroglyph datasets\footnote{Dataset available at: \url{https://huggingface.co/datasets/HamdiJr/Egyptian_hieroglyphs}} exhibit severe imbalance, where a few samples cover a large number of character classes. For example, the OBI dataset includes 2,806 distinct characters but only 12,170 samples, showing a pronounced long-tail distribution. This imbalance stems not only from historical losses but also from the domain-specific usage of ancient scripts \cite{liu2022one}, which often recorded specialized contexts like religious or ceremonial events, resulting in narrow vocabularies. These factors severely limit the availability of sufficient and diverse data, posing major challenges for recognition methods, particularly for data-driven approaches like machine learning, which rely on large, balanced datasets for optimal performance \cite{diao2023rzcr}.

Compared to general text images, ancient script images are notably affected by degradation. This degradation arises from the artifacts' physical condition, preservation environment, and historical copying processes \cite{krithiga2023ancient}, often resulting in noise that obscures large portions of text and critical character features, making recognition highly challenging \cite{gangamma2012restoration}. Moreover, due to the long preservation period, the noise is typically heterogeneous, combining various types and levels \cite{nafchi2013application}, such as scratches, stains, erosion, and digitization artifacts, further complicating character recognition \cite{zhao2020improvement}. One strategy is to preprocess images with denoising algorithms before recognition, but such methods have limitations. Unlike synthetic noise, which can be simulated and controlled, real ancient script noise is unpredictable and complex \cite{shi2022rcrn}. Fig.~\ref{img3} (a)-(d) shows examples of noisy ancient images, including Greek inscriptions, Egyptian hieroglyphs, Sumerian cuneiform, and OBIs, highlighting the prevalence of mixed noise types. Fig.~\ref{img3} (e) contrasts an OBI image overlaid with synthetic Gaussian noise, revealing significant visual differences and underscoring the challenge real-world noise poses, thereby reducing the effectiveness of models trained only on synthetic noise.

The challenges of limited sample availability and the presence of degradation highlight the unique difficulties in recognizing ancient scripts. These observations underscore the necessity of developing specialized recognition algorithms to address the distinct challenges posed by ancient scripts. This also forms one of the key motivations for this survey, which aims to organize recognition and processing methods tailored to the specific issues inherent in ancient script images.

% \subsection{Positioning and Contributions}
% \subsection{Our Survey in Context}
\subsection{Distinction from Prior Work}

\begin{table}[!t]
\centering
\resizebox{0.95\textwidth}{!}{%
\begin{tabular}{cccccc}
\toprule
\makecell{\textbf{Surveys}} &
\makecell{\textbf{Year} \\ \textbf{Published}} &
\makecell{\textbf{Ancient} \\ \textbf{Scripts}} &
\makecell{\textbf{Coverage} \\ \textbf{of Scripts}} &
\makecell{\textbf{Methods for} \\ \textbf{Imbalanced Data}} &
\makecell{\textbf{Methods for} \\ \textbf{Image Degradation}} \\
\midrule
Shah and Badgujar 
\cite{shah2013devnagari} & 2013 & \checkmark & Devnagari scripts& & \checkmark  \\
Dineshkumar and Suganthi 
\cite{dineshkumar2013research} & 2013 & \checkmark & Sanskrit scripts& & \\
Kataria and Jethva 
\cite{kataria2019cnn} & 2017 & \checkmark & Arabic, Latin, Sanskrit & &  \\
Djaghbellou et al. 
\cite{djaghbellou2021survey} & 2021 & & Arabic scripts & &  \\
Hirugade et al. 
\cite{hirugade2022survey} & 2022 & & Devnagari scripts & &  \\
Krithiga et al. 
\cite{krithiga2023ancient} & 2023 & \checkmark & Tamil scripts&  \\
Shen et al. 
\cite{shen2023challenges} & 2023 & \checkmark & Chinese scripts& &  \\
Chirimilla and Vardhan 
\cite{chirimilla2022survey} & 2022 & & English, Indian & & \checkmark  \\
% Naseer et al. \cite{naseer2024investigating} & 2024 & & Arabic, Hindi, Chinese and English & & \checkmark & \\
Narang et al. 
\cite{narang2020ancient} & 2020 & \checkmark & 12 scripts without taxonomy & &  \\
Al-Taee et al. 
\cite{al2020handwritten} & 2020 & & 7 scripts without taxonomy & &  \\
\textbf{Ours} & 2025 & \checkmark & \textbf{17} scripts \textbf{with} taxonomy & \checkmark & \checkmark \\
\bottomrule
\end{tabular}%
}
\caption{Comparison of prior surveys for ancient script recognition methods based on coverage of scripts, taxonomy, datasets provided, and technical challenges.}
\label{tab:script_survey_comparison}
\end{table}

Our main contribution is to offer, for the first time, a comprehensive survey that compares all existing ones on ancient script recognition. We demonstrate the distinction from existing surveys and reviews in Table~\ref{tab:script_survey_comparison}. As can be observed, current surveys that specifically address ancient script recognition remain limited in number, and many are relatively outdated. Moreover, most existing surveys focus on a narrow subset of scripts or overlook the importance of script taxonomy in method analysis. This leads to a lack of connection between the inherent features of different scripts, limiting further discussions on recognition techniques. Furthermore, we observe that the unique challenges posed by ancient scripts are often underexplored in previous works, with little emphasis on the technical solutions needed to address them. In summary, the following key points outline the advancements of this survey compared to existing studies:
\begin{enumerate}
    \item Broader Coverage with Structured Taxonomy: We cover a wider range of ancient scripts and employ an efficient taxonomy to present recognition methods for different scripts;
    \item Discussion on Data Imbalance: We discuss the challenge of data imbalance in ancient scripts and review corresponding learning methods and strategies;
    \item Analysis of Degraded Images: We analyze recognition and processing methods for degraded ancient script images.
    \item Up-to-date Methods: We include more up-to-date computer vision methods specific for ancient scripts recognition;
\end{enumerate}

\section{Logographic Scripts}
\label{sec:Logographic}
\begin{figure}[!t]
  \centering
  \includegraphics[width=1\linewidth]{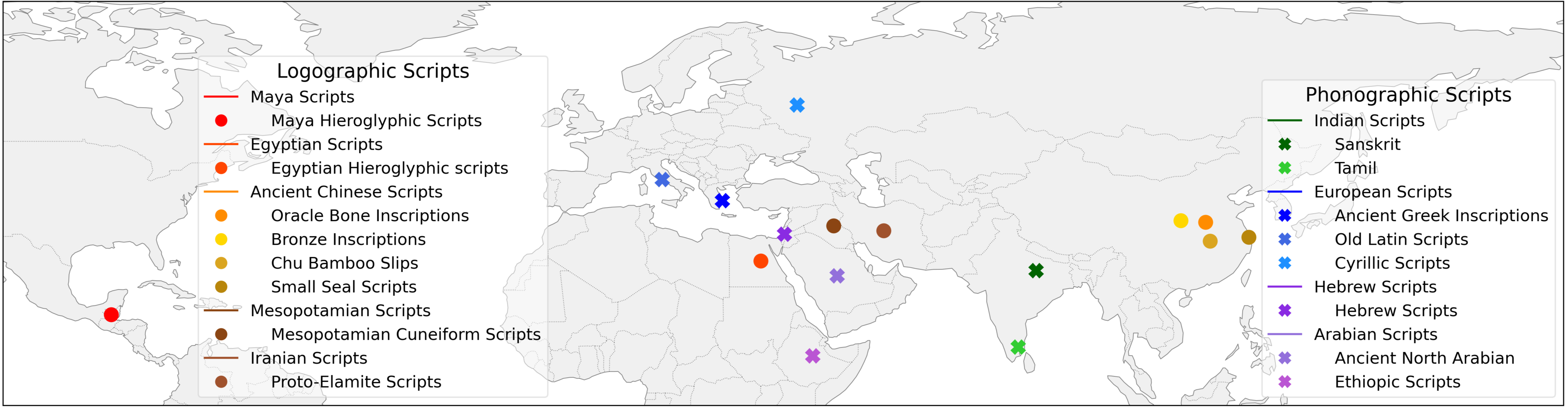}
  \caption{Geographic distribution of major ancient scripts. Logographic scripts are represented with colored circles, and phonographic scripts with colored crosses. Each script family shares a color hue, with script subtypes shown in varying shades. Marker positions reflect approximate historical usage regions.}
  \label{fig:Geographic_scripts}
\end{figure}

We categorize the ancient scripts into logographic and phonographic scripts. Building upon the linguistic family and geographic distribution of each script's origin and usage, we further organize and classify them in a more fine-grained manner. Fig.~\ref{fig:Geographic_scripts} presents the classification of all scripts along with their corresponding geographical distribution. The classification is grounded in the premise that the origin and usage of a script significantly influence the script's evolution. Scripts that share a common origin or that were historically used in geographically adjacent regions tend to exhibit similar glyphs and writing features. Such a systematic organization provides a foundation for a structured exploration of the technical challenges and solutions associated with each script family.

This section covers major ancient logographic scripts, including the Maya hieroglyphic scripts, ancient Egyptian hieroglyphic scripts, ancient Chinese scripts (oracle bone inscriptions, bronze inscriptions, Chu bamboo slips, and small seal scripts), ancient Mesopotamian cuneiform scripts, and the Proto-Elamite scripts from ancient Iran. These scripts, although diverse in geographic origin and stylistic form, share the fundamental features of combining visual symbolism with semantic encoding. Detailed descriptions of each script's writing characteristics and historical context are centralized here, whereas the subsequent subsections focus primarily on specific recognition technologies, dataset development, and modeling strategies relevant to each script.

\subsection{Maya Hieroglyphic Scripts}

% \begin{figure}[!t]
%   \centering
%   \includegraphics[width=0.65\linewidth]{images/Maya1.png}
%   \caption{Examples of Maya scripts.
%     \label{img:Maya}}
% \end{figure}

\begin{figure}[!t]
  \centering
  \includegraphics[width=1\linewidth]{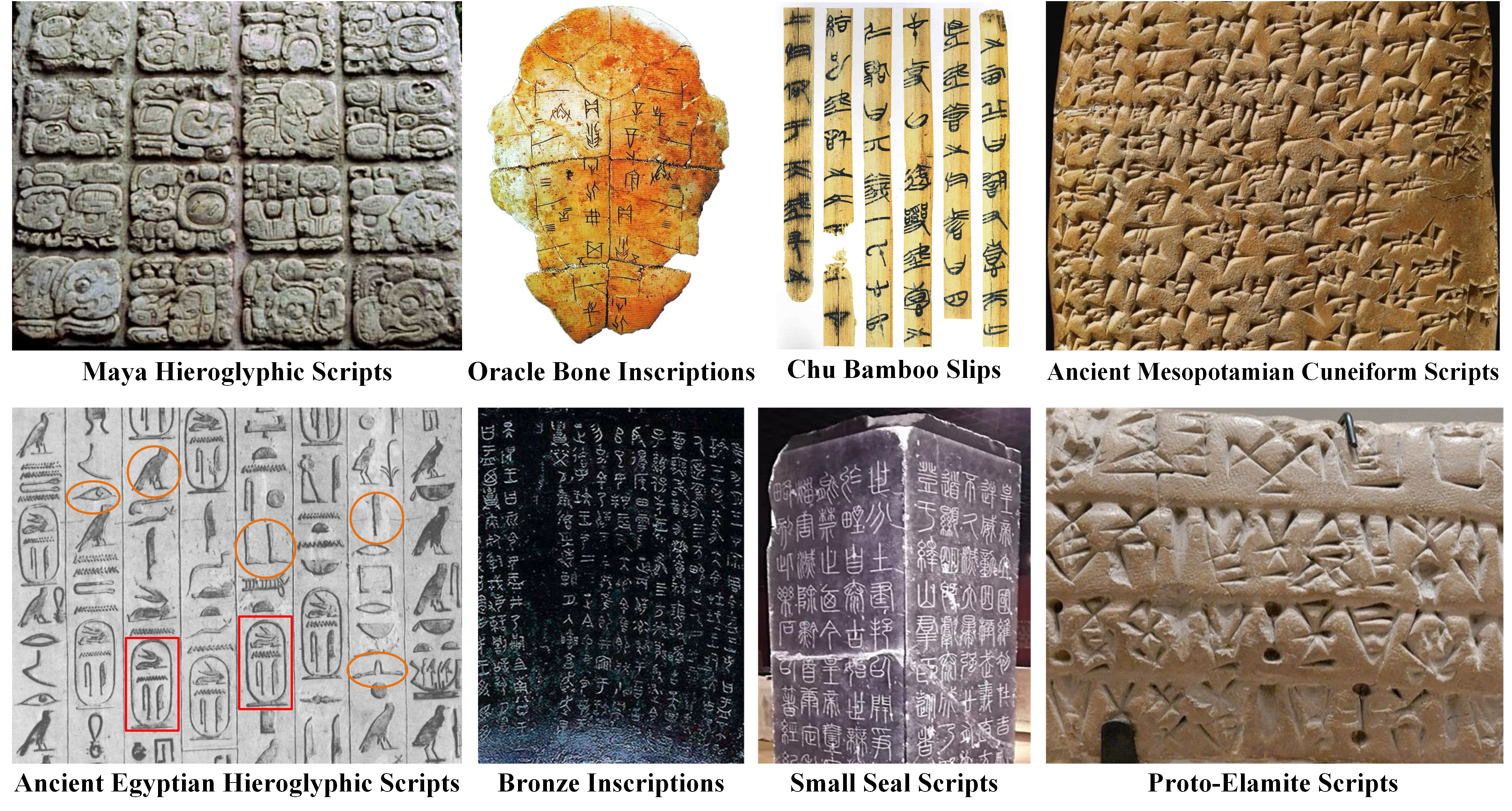}
  \caption{Examples of logographic scripts from diverse ancient civilizations, including Maya hieroglyphs, Egyptian hieroglyphs, Oracle Bone inscriptions, Bronze inscriptions, Chu bamboo slips, Small Seal script, Mesopotamian cuneiform, and Proto-Elamite script. These scripts were inscribed on various writing media, e.g., stone, bone, bronze, bamboo, and clay. In the Ancient Egyptian panel, red boxes highlight royal names enclosed in cartouches, while orange circles indicate characters derived from real-world objects and animals.}
    \label{fig:logographic}
\end{figure}

Maya hieroglyphic scripts constitute a highly pictorial writing system that was extensively used to document core aspects of ancient Maya civilization (300 BCE - 900 CE), including religious ceremonies, political affairs, and astronomical observations. These glyphs are distributed across present-day Mexico, Guatemala, and Honduras. As shown in Fig.~\ref{fig:logographic}, characterized by a highly flexible graphemic structure and diverse writing orientations, Maya hieroglyphs pose considerable challenges for computational recognition and digitization. Early studies in this domain primarily focused on image retrieval. For instance, Roman et al.\ initiated the AJIMAYA project \cite{roman2009retrieving}, which brought together archaeologists and computer scientists in a cross-disciplinary collaboration. The project collected high-resolution images and hand-drawn sketches from several UNESCO World Heritage sites and conducted the first systematic evaluation of shape context descriptors for the recognition of syllabic Maya glyphs. Building on this foundation, Hu et al.\ proposed a glyph retrieval approach that combined shape similarity with contextual information \cite{hu2014automatic}. By modeling adjacent glyphs as a first-order Markov chain, their method enabled re-ranking of retrieval results based on co-occurrence probabilities, significantly improving the accuracy of visual matching.

To address the structural and stylistic complexity of Maya hieroglyphs, researchers explored interactive and visualization-based strategies to enhance the operability of image analysis workflows. In a subsequent study, Roman et al. expanded their glyph image dataset to include over 3,400 characters and introduced refinements to the HOOSC descriptor, making it more robust against open contours, variable stroke thicknesses, and intricate internal ornamentation. These improvements led to a 20\% gain in retrieval performance \cite{roman2011searching}. Additionally, they analyzed statistical models for glyph classification \cite{roman2011analyzing}, further investigating co-occurrence statistics at the linguistic level to support the quantitative interpretation of ancient texts. Emphasizing the role of human-computer interaction, Hu et al. \cite{hu2017analyzing} proposed a graph-based visualization interface in which nodes represent individual glyph images and edges encode visual similarity. This tool provided an intuitive means for scholars to explore complex hieroglyphic datasets and highlighted the value of human-in-the-loop approaches in ancient script recognition. In parallel, Bogacz et al. \cite{bogacz2018visualizing} developed an image analysis pipeline that integrates multi-scale invariant filtering, random walk segmentation, and histogram of oriented gradients (HOG)-based clustering. Their method uncovered latent relationships among previously unidentified glyphs, offering new insights into the connections between visual style and linguistic structure.

With the advancement of deep learning, convolutional neural networks (CNNs) have been introduced into Maya glyph recognition tasks to address the limitations of earlier techniques in handling the script's visual complexity and variability. Can et al. \cite{can2018tell} conducted a systematic comparison of different CNN training strategies, including transfer learning, fine-tuning, and training from scratch. Their experiments across several architectures provided insights into the trade-offs between model depth and performance. They also employed GradCAM-based visualization to enhance the interpretability of learned representations, offering domain experts a more transparent view into the model's decision-making process. 
Roman et al.\ \cite{roman2016transferring} further utilized intermediate-layer representations from the VGG network for glyph indexing and visualization. By applying dimensionality reduction techniques, they improved dataset navigability while emphasizing the inherent spatial flexibility of Maya script, such as glyph resizing and positional rearrangement, as a key challenge for computational modeling. More recently, Fnu et al.\ addressed the issue of segmentation accuracy by proposing a fine-tuned segmentation pipeline based on pre-trained models \cite{fnu2024segmentation}. Developed in collaboration with experts in Maya art and history, their method achieved promising performance in glyph boundary detection.

\subsection{Ancient Egyptian Hieroglyphic Scripts}

% \begin{figure}[!t]
%   \centering
%   \includegraphics[width=0.65\linewidth]{images/egyptian.jpg}
%   \caption{Examples of Ancient Egyptian Hieroglyphic Scripts.
%     \label{img:Egyptian}}
% \end{figure}

Ancient Egyptian hieroglyphic scripts represent one of the most intricate writing systems in human history, with their use spanning from approximately 3,100 BCE - 500 CE. These pictorial symbols, which are grounded in real-world objects, animals, humans, or abstract forms, were primarily used in religious inscriptions and official documents, as shown in Fig.~\ref{fig:logographic} with orange circles. Early explorations into the automated recognition of hieroglyphs focused on classical machine learning techniques and pattern-matching approaches. Franken et al. proposed a recognition system that combined image processing with statistical language models, utilizing a dataset of 3,993 annotated glyphs from the Pyramid of Unas. Their work underscored the importance of integrating visual features with semantic context and marked an early effort in combining visual and linguistic models for hieroglyphic recognition \cite{franken2013automatic}. Duque et al. concentrated on the recognition of cartouches\footnote{An oval frame enclosing royal names} as shown in Fig.~\ref{fig:logographic}, and introduced a three-stage image processing pipeline involving shape matching, edge detection, and Levenshtein distance. Their method demonstrated strong performance in name identification and historical correlation, laying a practical foundation for applications such as museum-guided systems \cite{duque2017deciphering}.

Given the pictographic nature and stylistic diversity of hieroglyphic forms, conventional pattern-matching techniques often fall short. This has increasingly led to the adoption of deep learning approaches in the study of ancient scripts. Barucci et al.\ introduced significant advancements in hieroglyph recognition using deep learning algorithms. Their Glyphnet model integrates and optimizes several CNN architectures, including ResNet-50, Inception-v3, and Xception, achieving promising classification performance \cite{barucci2021deep}. They further extended their work to address the task of glyph segmentation, thereby enabling higher-level computational analyses such as symbol recognition, phonetic transliteration, and broader Egyptological applications \cite{barucci2022ancient}. In a parallel effort, Guidi et al.\ leveraged the Detectron2 framework to propose a deep learning-based instance-level segmentation method. As one of the first systematic studies focused on symbol-level instance recognition, their model, based on a fine-tuned Mask R-CNN, achieved robust segmentation performance across heterogeneous image sources, providing a precise foundation for transcription and stylistic analysis under complex visual conditions \cite{guidi2023egyptian}.

Concurrently, scholars have expanded their focus to the semantic interpretation of hieroglyphic images, aiming to extract the underlying meanings encoded within the script. Elnabawy et al.\ proposed a comprehensive recognition pipeline based on image processing and optical character recognition (OCR), encompassing image acquisition, automatic segmentation, reading order control, post-processing, and HOG feature matching. Their system integrated the Gardiner sign list to significantly enhance automation and reading sequence accuracy, and stands as an early full-stack system aimed at ``image-to-interpretation" tasks \cite{elnabawy2018image}. In a more interactive direction, Plecher et al.\ developed ARsinoë, an augmented reality system that merges long short-term memory (LSTM) networks and 3D interaction design to support recognition and dynamic interaction with both printed and handwritten hieroglyphs \cite{Plecher2020Hieroglyphs}. Further advancing this line of research, Sobhy et al.\ introduced an end-to-end hieroglyph interpretation framework that integrates R-CNN-based object detection, Siamese networks for few-shot classification, and character-level language modeling \cite{Sobhy2023Hieroglyphic}.

\subsection{Ancient Chinese Scripts}

% \begin{figure}[!t]
%   \centering
%   \includegraphics[width=0.95\linewidth]{images/Ancient_Chinese1.png}
%   \caption{Examples of Ancient Chinese Scripts. (a) Representative artifacts and inscription media for four major ancient Chinese scripts: Oracle Bone Inscriptions, Bronze Inscriptions, Chu Bamboo Scripts, and Small Seal Scripts. (b) Examples of character forms as written in each script.
%     \label{img:Chinese}}
% \end{figure}

\subsubsection{Oracle Bone Inscriptions}
Oracle Bone Inscription (OBI) is the earliest known form of Chinese writing, primarily used during the Shang dynasty (around 1,300 BCE) for divination and ritual inscriptions. OBIs were typically carved on turtle plastrons and ox scapulae, as shown in Fig.~\ref{fig:logographic}. Due to centuries of weathering, fragmentation, and erosion, automatic recognition of OBIs presents unique technical challenges. Early studies primarily relied on image processing techniques and handcrafted feature extraction for character enhancement and recognition. Meng et al. \cite{meng2017recognition} employed a combination of Hough Transform, Gaussian filtering, and image binarization to effectively separate characters from background noise, reducing false positives and improving image clarity and subsequent recognition accuracy. Sun et al. \cite{sun2020dual} applied the scale-invariant feature transform (SIFT) algorithm to extract robust keypoints, enhancing character localization even in severely damaged inscriptions. By integrating a structured character database, their method achieved high recognition accuracy for incomplete glyphs. Chen et al. \cite{chen2020study} proposed an encoding-based recognition framework that transforms OBI images into compact matrices for efficient character identification. Their approach demonstrated high processing speed and accuracy on inscriptions from the Yin Ruins dataset.

In recent years, the advancement of image processing, deep learning, and generative modeling has significantly improved both recognition accuracy and task complexity for OBIs \cite{diao2025Oracle, yue2025Ancient}. Meng et al. \cite{meng2018recognition} were among the first to employ deep CNNs to extract visual features from OBIs. They applied data augmentation strategies, including rotation and brightness variation, to simulate real-world scenarios, achieving a recognition accuracy of 92.3\%. Xing et al. \cite{xing2019oracle} conducted a comparative evaluation of several object detection architectures. YOLOv3 was selected as the optimal baseline, and further optimized through anchor box adjustments and loss function modifications tailored to the characteristics of oracle bone rubbings. To address issues of class imbalance and limited samples, Zhang et al. \cite{zhang2019oracle} proposed a metric-learning-based CNN framework, using a triplet loss function to embed characters into a Euclidean space. This improved classification performance for rare and previously unseen character categories. At the same time, Huang et al. \cite{huang2019obc306} introduced the OBC306 dataset, which includes 309,551 samples across 306 OBI categories. This dataset has become a key benchmark for evaluating the performance of deep learning models. While architectures such as AlexNet and ResNet50 showed strong baseline performance, the study also highlighted the continued challenges of fine-grained classification.

In addition, Liu et al. \cite{liu2020oracle} investigated the use of lightweight CNN architectures such as SqueezeNet, which offered a favorable trade-off between computational efficiency and recognition performance. By applying extensive data augmentation, their model demonstrated enhanced accuracy in recognizing incomplete characters. For the task of OBI fragment reassembly, Zhang et al. \cite{zhang2021ai} introduced the OB-Rejoin dataset and developed a Siamese network-based method for fragment matching, achieving a reassembly accuracy of 80.9\%, thereby extending the application of OBI recognition into the domain of physical restoration. In terms of script variant clustering, Liu et al. \cite{liu2021recognition} combined ResNet50 feature extraction with spectral clustering to automatically group visually similar glyph variants. Guo et al. \cite{guo2022improved} proposed enhancements to the Inception-V3 architecture, incorporating rotation and scaling augmentations to improve robustness against noisy images. Wang et al. \cite{wang2022unsupervised} introduced the structure-texture separation network (STSN), which disentangles script structure from background noise. Their method demonstrated strong robustness in aligning printed and scanned OBIs, particularly under severe degradation. Xie et al. \cite{xie2024diffobi} explored the use of diffusion models and proposed DiffOBI, a conditional diffusion-based framework for oracle bone script image generation. By introducing a dual-branch style control module and a hierarchical conditional modeling strategy, the method effectively generates high-quality oracle-style glyphs with strong stylistic consistency, enhancing the performance of data augmentation in few-shot oracle character recognition tasks.

\subsubsection{Bronze Inscriptions}
Bronze inscriptions, also known as Jinwen, represent a significant form of ancient Chinese writing. They were widely used from the Western Zhou period (around 700 BCE) through to the Spring and Autumn and Warring States periods, commonly cast or engraved on ritual bronzeware such as ceremonial vessels and bells, as shown in Fig.~\ref{fig:logographic}. Compared to oracle bone script, bronze inscriptions feature more rounded and robust strokes with a greater diversity in structure. However, due to centuries of burial, the surfaces of these artifacts often exhibit oxidation, corrosion, and wear, which pose substantial challenges for image recognition and feature extraction. To address the recognition of bronze inscriptions, Wu et al. \cite{wu2022cnn} proposed a recognition method based on a backpropagation neural network tailored for bronze image classification. Their study constructed a training dataset comprising 100 distinct bronze images, each associated with 10 samples, and an independent test set containing 256 samples. The model relied on a feature extraction strategy that fused pixel-level information with local features, ultimately achieving a recognition accuracy of 93.3\%. This result demonstrates the potential of deep neural networks in handling visually complex bronze images with noisy backgrounds. Wang et al. \cite{wang2020bronze} proposed a more advanced convolutional neural network (CNN)-based model for the recognition of bronze inscriptions. Their work systematically compared and optimized multiple CNN architectures to adapt to the unique morphological characteristics and surface variability of bronze inscriptions. The model design incorporated a spatial transformer network (STN) to enhance focus on key inscription regions and employed Implicit Semantic Data Augmentation to improve the model's generalizability and robustness. Evaluated on a large-scale bronze inscription dataset, the final model achieved an accuracy of 91.2\%, validating the effectiveness and scalability of deep learning approaches in the interpretation of complex historical scripts.

\subsubsection{Chu Bamboo Slips}
During the Warring States period, bamboo slips were widely used in the state of Chu (around 400 BCE) to document political, philosophical, and historical texts. As shown in Fig.~\ref{fig:logographic}, these texts were inscribed on thin, narrow strips of bamboo using brush and ink, a lightweight and flexible medium that preceded the invention of paper. Chu slip characters feature fluid strokes and loose structures, making them more challenging to recognize than carved scripts. To better capture such stylistic complexity, Wu et al. \cite{wu2021ancient} proposed a transformer-based recognition model incorporating multi-scale attention and MLP modules, along with graphical augmentation to improve training stability and convergence. Recognition tasks involving Chu slips often encounter issues such as stroke overlap, uneven ink distribution, and background noise. Addressing character segmentation under these conditions, Cao et al. \cite{cao2022character} introduced a locally adaptive thresholding method, using an Effective Character Contour Length (ECCL) metric and multi-Gaussian fitting to determine optimal thresholds. Focusing on text detection, Jing et al. \cite{jing2024method} proposed an enhanced deep learning framework that integrates an IntraCL module, an adaptive feature pyramid, and a refined loss function, significantly improving detection accuracy in noisy and densely written slips.

\subsubsection{Small Seal Scripts}

Small seal script, also known as Zhuanshu, was the standardized script promoted during the Qin dynasty (around 200 BCE). It is characterized by rounded strokes, highly regular yet intricate structures, as shown in Fig.~\ref{fig:logographic}. Xu et al. \cite{xuchinese} constructed the first large-scale Chinese seal dataset, which includes seals of various colors and difficulty levels. They developed a complete pipeline for detection, segmentation, and recognition, and established baseline models using YOLOv3 and related architectures, thereby extending the paradigm of traditional OCR research. Some studies focused on low-resource scenarios. Roy et al. \cite{roy2009seal} proposed a structural hashing method that leverages spatial relationships between character pairs. Through hash table indexing and a voting mechanism, their approach achieved rotation and scale-invariant recognition, even when characters were partially missing, demonstrating robustness in complex visual conditions. Ou et al. \cite{ou2024qin} combined HOG features with an SVM classifier, using Gaussian filtering and gamma correction to reduce noise.

Despite these advancements, the complexity of small seal script continues to pose significant challenges for recognition and restoration. To address these complexities, Tang et al. \cite{tang2023ancient} developed a lightweight CNN, ShuiNet-A, which integrates attention mechanisms to enhance the modeling of intricate stroke patterns. Wenjun et al. \cite{wenjun2023ea} proposed EA-GAN, a restoration framework based on GANs, to recover missing small seal scripts in historical works. The model employs a dual-branch architecture to extract multi-scale contextual features and uses reference images to guide generation, effectively simulating human handwriting restoration and outperforming traditional interpolation-based methods. Zhou et al. \cite{zhou2023style} addressed the zero-shot recognition by introducing a style-independent pictographic radical sequence learning framework. The model aligns small seal and traditional Chinese characters using GAN-based style normalization and employs a transformer to learn radical sequences. It demonstrated strong performance across three test sets, particularly in recognizing unseen characters and enabling cross-style translation.

\subsection{Ancient Mesopotamian Cuneiform Scripts}
Cuneiform is one of the earliest known writing systems in human history (around 3000-600 BCE), originating in the region of Mesopotamia (present-day Iraq and parts of Syria) and remaining in use for over three millennia. It was typically impressed onto wet clay tablets using a reed stylus, producing wedge-shaped marks with well-defined strokes, as shown in Fig.~\ref{fig:logographic}. Cuneiform is characterized by its strong directionality, abstract structural composition, and large inventory of symbols. In early computational studies, Bogacz et al. \cite{bogacz2015character} proposed a wedge-based recognition method by extracting geometric feature vectors of individual wedges and reformulating character recognition as a linear assignment optimization task, enabling robust retrieval without word boundaries or annotated data. Mostofi et al. \cite{mostofi2014intelligent} developed a recognition system for old Persian cuneiform that integrates image preprocessing with a feedforward neural network, incorporating Gaussian noise simulation to improve robustness under degraded image conditions and achieving 89\% accuracy across 46 character classes. Kriege et al. \cite{kriege2018recognizing} introduced two graph-based recognition strategies: one leveraging a graph edit distance function based on vertex types, directional edges, and spatial relationships; the other employing a graph convolutional network (GAN) for direct and efficient feature extraction and classification on structural graphs.

To address the challenges posed by complex character structures and limited annotated data, Mara et al. constructed the HeiCuBeDa dataset \cite{mara2019breaking}, which comprises 3D models of 707 clay tablets with high-resolution surface features, six-view raster images, transliterations, phonetic interpretations, and metadata. This resource significantly lowers the barrier for applying computational methods to cuneiform analysis. Stotzner et al. \cite{stotzner2023cnn} proposed a CNN-based framework for symbol detection and classification and introduced an annotation mapping mechanism between rendered images and original photographs, confirming the utility of synthetic rendering for symbol localization and recognition. ElShehaby et al. \cite{elshehaby2024unlocking} employed transfer learning using a VGG16-based model to develop a cuneiform symbol recognition system focused on scanned archaeological artifacts, particularly symbols from the Code of Hammurabi. With data augmentation, the system achieved competitive performance in symbol recognition.

In recent years, research on cuneiform recognition has gradually shifted from traditional image-based methods to multimodal fusion and interactive human-in-the-loop systems. Mousavi et al. \cite{mousavi2017extracting} proposed an integrated system that combines recognition, pronunciation, and translation, using preprocessing techniques such as median filtering, morphological operations, and edge blurring. The system achieved 92\% recognition accuracy on both handwritten and inscribed texts and introduced a database of 1,500 annotated samples for old Persian cuneiform. Gordin et al. \cite{gordin2022optical} developed CuRe, an interactive platform supporting multi-stage recognition of hand-copied cuneiform, including character detection via Faster R-CNN, symbol classification with ResNet18, and a stroke-level segmentation module. Additionally, Bogacz et al. \cite{bogacz2022digital} provided a systematic review of recognition techniques across 2D images, vector graphics, and 3D scans, evaluating the applicability of curvature descriptors, keypoint-based modeling, GNNs, and GANs in cuneiform modeling workflows. Hagelskjaer et al. \cite{hagelskjaer2022deep} further introduced a large-scale point cloud network architecture tailored for classifying metadata of cuneiform tablets and proposed a maximum attention visualization method, enhancing model interpretability and achieving state-of-the-art results on benchmark datasets.

\subsection{Proto-Elamite Scripts}
Proto-Elamite, one of the earliest known writing systems in ancient Iran, dates back to approximately 3100 BCE. This script combines pictographic and abstract symbols, with a consistent linear arrangement. As shown in Fig.~\ref{fig:logographic}, Proto-Elamite is considered a variant of cuneiform writing, but it differs notably from contemporaneous Mesopotamian cuneiform in several respects. While both employ wedge-shaped impressions, Proto-Elamite features more abstract, often less standardized symbols and exhibits a greater emphasis on alignment and tabular organization, especially in numerical and accounting contexts. Due to the substantial structural variability and frequent graphical mutations of its characters, a significant portion of the script remains undeciphered, posing considerable challenges for scholarly research. At the writing system level, Englund \cite{englund2001state}, under the framework of the Cuneiform Digital Library Initiative, systematically examined the structural characteristics of Proto-Elamite in terms of format, semantic organization, and numerical systems. He highlighted its higher degree of linearity relative to contemporary proto-cuneiform, along with its use of more static and diversified counting methods. Dahl \cite{dahl2018proto}, meanwhile, analyzed Proto-Elamite from the perspectives of script origin, regional distribution, and grammatical structure. Through stratigraphic analysis and internal textual features, Dahl also established a relative chronology for Proto-Elamite documentation.

In recent years, researchers have increasingly turned to computational approaches to analyze the structure of compositional graphemes within Proto-Elamite in order to uncover their combinatorial patterns and potential semantic rules. Born et al. \cite{born2021compositionality} employed a multimodal language model that jointly processes image sequences and associated labels, revealing that Proto-Elamite symbols exhibit a degree of compositional regularity. Their study demonstrated that image-driven models offer superior interpretability compared to text-only methods, thereby reducing annotator bias and uncovering latent structural patterns. Extending this line of work, Born et al. \cite{born2023applications} introduced a multimodal representation learning framework that integrates image encoders with language models. They also proposed a bootstrapped classification algorithm and sequence modeling strategies to advance structural understanding across multiple dimensions, including character clustering, composite grapheme construction, numeric symbol disambiguation, and document structure analysis. These studies indicate that while Proto-Elamite graphemes exhibit a high degree of graphical freedom, their internal organization is nonetheless systematic, providing a technical pathway for deciphering this yet-untranslated script. In the DeepScribe project, Williams et al. \cite{williams2023deepscribe} implemented a deep learning pipeline combining RetinaNet for symbol localization and ResNet for symbol classification. This work achieved high accuracy in both detection and classification tasks, demonstrating the feasibility of automated transcription for ancient cuneiform scripts.

\section{Phonographic Scripts}
\label{sec:Phonographi}
We cover nine major ancient phonographic scripts here: Sanskrit (including Devanagari), Tamil, ancient Greek inscriptions, Old Latin, Cyrillic, Hebrew, ancient North Arabian, and Ethiopic script. In view of their linguistic origin and historical usage as shown in Fig.~\ref{fig:Geographic_scripts}, they are classified into several subcategories: ancient Indian scripts, European alphabetic scripts, Hebrew script, and ancient Arabian scripts. Our aim is to reflect the historical transmission and evolution of the phonographic scripts' tradition, and to reorganize them according to their developmental trajectory. Through this perspective, we will uncover shared patterns and methodological divergences in their recognition methods. We provide a detailed overview of the characteristics of these scripts, with the subsequent subsections focusing on computational approaches and recognition research specific to each script.

\begin{figure}[!t]
  \centering
  \includegraphics[width=1\linewidth]{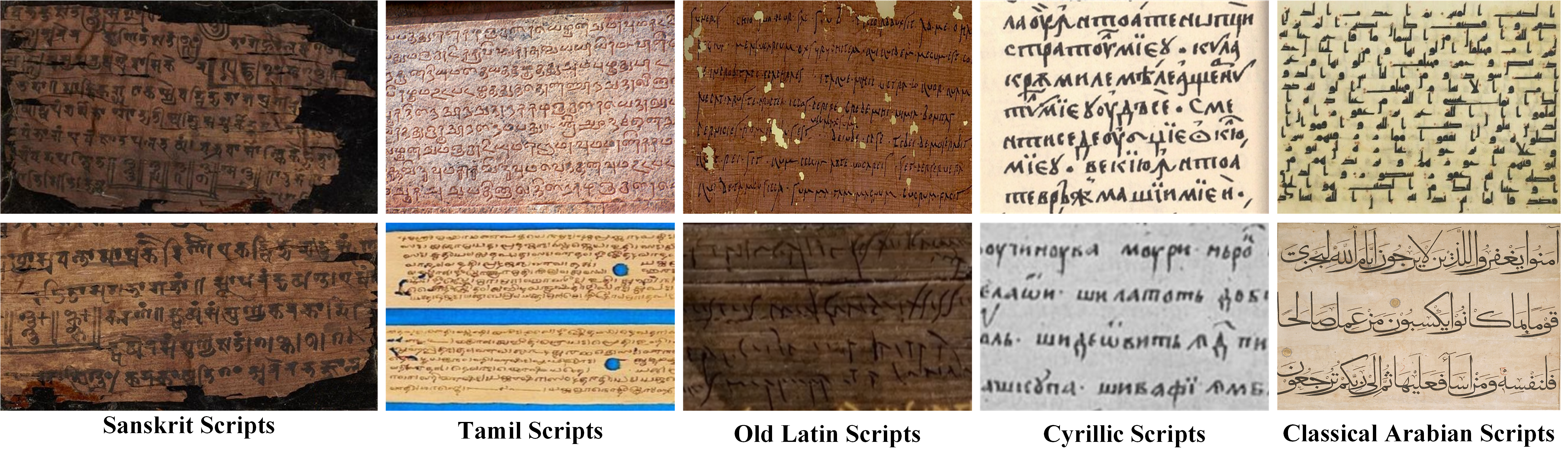}
  \caption{Complex ligature patterns are a defining feature across Sanskrit, Tamil, Old Latin, Cyrillic, and Classical Arabian scripts, and appear extensively in their manuscript traditions. These structures introduce shared challenges in OCR, often requiring similar recognition strategies across languages.}
    \label{fig:ligature}
\end{figure}

\subsection{Ancient Indian script}

\subsubsection{Sanskrit Script}
Sanskrit is one of the oldest and most culturally authoritative languages of the Indian subcontinent. It was originally written in the Brahmi script (around 300 BCE) and later evolved into various writing systems, including Devanagari (around 700 CE). As shown in Fig.~\ref{fig:ligature}, due to its complex ligature structures, the Sanskrit script often exhibits non-linear visual configurations, leading to blurred character boundaries and increased recognition difficulty. To address these challenges, researchers have proposed various frameworks based on both traditional approaches and deep learning techniques for high-precision character- or word-level recognition. In earlier studies, Dwivedi et al. proposed a recognition framework that integrates edge detection, genetic algorithms, and SVM classifiers to tackle issues related to structural redundancy and segmentation ambiguity in handwritten Sanskrit scripts \cite{dwivedi2013sanskrit}. Dineshkumar et al. developed an offline OCR system based on a feedforward neural network that encompasses the full pipeline from image preprocessing and character segmentation to feature extraction and classification \cite{dineshkumar2015sanskrit}. Shih et al. designed a two-stage classification system that employs k-nearest neighbors for coarse classification, followed by multi-class SVMs for refined recognition, achieving an accuracy of 85.6\% across 65 classes of handwritten Sanskrit scripts \cite{shih2011machine}.

The diverse font variants and highly compact writing style of classical Sanskrit scripts introduce additional challenges to the recognition task. With the advancement of deep learning, researchers have increasingly adopted architectures such as CNNs, LSTMs, and attention mechanisms to enhance context understanding and sequential modeling. Dwivedi et al. proposed an OCR system that combines a CNN-based feature extractor with an attention-driven Bidirectional LSTM (Bi-LSTM) sequence model, specifically addressing the challenge of recognizing long compound ligatures in classical Sanskrit texts, supported by a large-scale hybrid dataset consisting of both real and synthetic samples \cite{dwivedi2020ocr}. Kataria et al. introduced a CNN-Bi-LSTM-based model for printed Sanskrit text recognition, achieving 98.6\% character-level and 93.5\% paragraph-level accuracy without relying on any language model, thereby significantly improving the system's practical applicability \cite{kataria2021optical}. Lomte et al., targeting handwritten Sanskrit, constructed a dataset of 70,000 images and proposed three variants of a four-fold CNN architecture, benchmarking against AlexNet \cite{lomte2022handwritten}. Kulkarni et al. demonstrated through comparative experiments that CNN-based methods offer superior performance in text-background separation and character boundary enhancement compared to traditional models such as KNN and SVM, making them well-suited for preprocessing tasks in Sanskrit manuscript recognition \cite{kulkarni2022proposed}. 

\noindent\textbf{Devanagari Script} is a widely used script in India and Nepal. The handwritten forms of Devanagari appear extensively in historical documents and exhibit significant stylistic differences compared to modern printed forms. These variations in glyph structure, along with the challenges of character connectivity and blurred boundaries, present considerable difficulties for character segmentation and recognition. Feature engineering methods continue to play a role in Devanagari character recognition. Narang et al. applied the discrete cosine transform to extract zigzag scan features from character images and combined them with an AdaBoost ensemble classifier to effectively recognize Devanagari characters in historical manuscripts \cite{narang2019devanagari}. To further model sequential dependencies in text, Moudgil et al. proposed a hybrid model combining CNN and LSTM architectures: CNNs are responsible for image-level feature extraction, while LSTMs capture temporal dependencies between characters. This method achieved an accuracy of 93.6\% in ancient manuscript recognition tasks \cite{moudgil2022cnn}. Building on this, Jindal et al. implemented a CNN-Bi-LSTM hybrid architecture with Bayesian optimization for automated hyperparameter tuning, reaching a recognition accuracy of 96.9\% for Sanskrit and 95.8\% for Maithili, demonstrating the robustness of such models in handling faded or stylized characters \cite{jindal2024hybrid}. Additionally, Moudgil et al. enhanced a dataset of over 7,000 character images by applying rotation and flipping augmentation techniques to 85 pages of Sanskrit manuscripts, and used a transfer learning model based on AlexNet to achieve reliable recognition performance \cite{moudgil2024devanagari}.

\subsubsection{Tamil Script}
Tamil is an important language spoken in southern India and Sri Lanka. Its writing system belongs to the abugida family, characterized by consonant-vowel combinations. Historically, it was used in palm-leaf manuscripts and stone inscriptions, and it originated around the 2nd century BCE. Tamil characters in handwritten script are tightly connected and include a large number of ligatures and character variants. Some examples are given in Fig.~\ref{fig:ligature}. Compared to typical phonographic scripts, Tamil has a significantly larger set of characters. These features collectively increase the complexity of automatic recognition. Kannan et al. proposed a complete character processing pipeline combining preprocessing, skew correction, feature extraction, and a Kohonen Self-Organizing Map classifier, achieving robust recognition across different handwriting styles \cite{kannan2009comparative}. Rajakumar et al. were the first to apply SVM to handwritten Tamil character recognition, using pixel-density-based region features and a multi-class SVM classification strategy, reaching an overall recognition accuracy of 82.0\% over 106 character classes \cite{rajakumar2011century}. Shanthi et al. conducted a comprehensive comparison of printed, handwritten, online, and offline OCR methods for Tamil script, and proposed a hybrid recognition framework integrating HMMs, SVMs, and neural networks \cite{shanthi2010novel}. Banumathi et al. further improved feature extraction and noise reduction by combining Contourlet transform and fuzzy median filtering, and applied neural networks to restore characters and estimate their historical dating \cite{banumathi2011handwritten}.

The diverse writing media of ancient Tamil script have introduced additional challenges for recognition tasks. In the domain of historical manuscript analysis, researchers have extended OCR efforts to more difficult materials such as stone inscriptions and palm-leaf manuscripts. Manigandan et al. proposed a method combining image preprocessing, SIFT feature extraction, and SVM classification to recognize characters in ancient Tamil stone inscriptions. They also integrated trigram-based language modeling to enhance recognition stability \cite{manigandan2017tamil}. Pragathi et al. developed a large-scale handwritten Tamil character dataset with 40,000 samples and introduced a CNN-based recognition system with multiple convolution and pooling layers as a base model \cite{pragathi2019handwritten}. For Tamil-Brahmi script recognition, Subadivya et al. applied a CNN model augmented with rotation and cropping techniques to address data scarcity, achieving a recognition accuracy of 94.6\% \cite{subadivya2020tamilbrahmi}. Furthermore, Eswaran et al. constructed a dedicated character dataset for palm-leaf manuscript recognition and proposed a hybrid feature extraction method combining triangular zoning and salient region slicing. When paired with an SVM classifier, the system achieved a recognition accuracy of 90.1\%, representing a 9.1\% improvement over the baseline models \cite{eswaran2021recognizing}.

\begin{figure}[!t]
  \centering
  \includegraphics[width=1\linewidth]{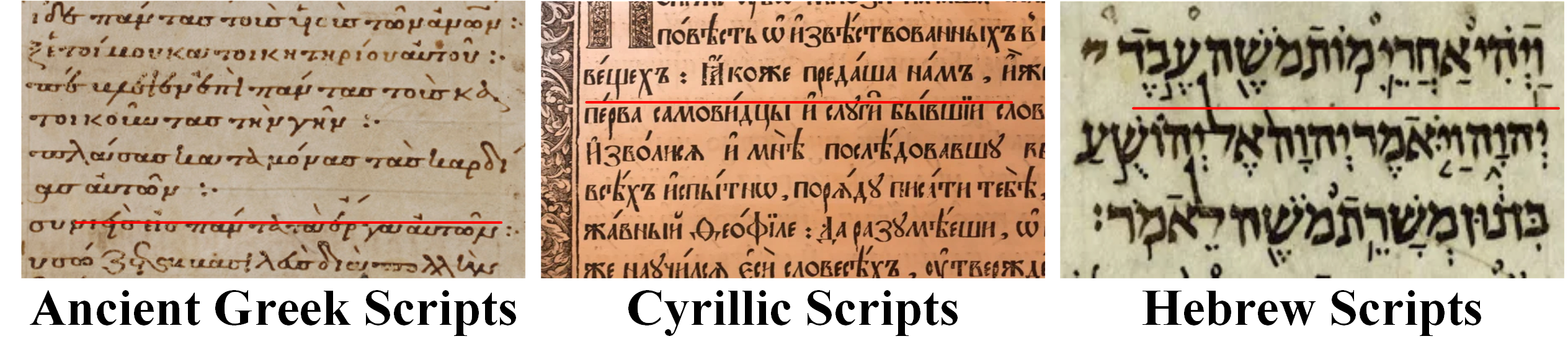}
  \caption{Examples of diacritical marks from Ancient Greek, Cyrillic, and Hebrew manuscripts, including breathing signs, accentuation, cantillation marks, and vocalization dots, contribute to the visual and typographic complexity of the scripts. Representative examples of diacritical marks are highlighted with red lines.}
    \label{fig:diacritical}
\end{figure}

\subsection{European Alphabetic Scripts}

\subsubsection{Ancient Greek Scripts}
Ancient Greek script, as one of the significant precursors to modern Western alphabetic systems, has long been a major focus for tasks involving recognition and interpretation. Ancient Greek script originated in the 8th century BCE. One of its notable features is the extensive use of diacritical marks, such as accent symbols, which significantly increase the number of possible character combinations and consequently add complexity to recognition tasks. Some examples of diacritical marks in Ancient Greek Scripts are shown in Fig.~\ref{fig:diacritical}. To address this challenge, Simistira et al. proposed a Bi-LSTM-based recognition system trained on a combination of real scanned documents and synthetic data, achieving substantial improvements in the recognition accuracy of polytonic Greek manuscripts \cite{simistira2015recognition}. Considering the variability introduced by ligatures and individual handwriting styles, Swindall et al. developed the AL-ALL and AL-PUB datasets. They compared the performance of various CNN architectures and demonstrated that ResNet exhibited stronger robustness when dealing with complex fonts \cite{swindall2021exploring}. Markou et al. proposed a CRNN-FCNN architecture combining CNNs and Bi-LSTM networks, and validated its superior performance in historical Greek manuscript recognition through a dataset named EPARCHOS \cite{markou2021convolutional}. To tackle challenges related to multilingual and complex document layouts, Romanello et al. introduced the GT4HistComment and Pogretra datasets, along with a novel OCR evaluation approach based on layout-region granularity \cite{romanello2021optical}. In addition, Tsochatzidis et al. enhanced handwritten text recognition (HTR) performance on Greek historical manuscripts by integrating Octave CNNs with Bidirectional GRUs (Bi-GRUs) \cite{tsochatzidis2021htr}, while Platanou et al. systematically investigated the influence of script style evolution across the 10th to 16th centuries and released the HPGTR dataset \cite{platanou2022handwritten}. For character detection and recognition, Swindall et al. used GAN-based data augmentation techniques to improve the accuracy of handwritten Greek characters recognition, and introduced a blockchain-based framework to manage transcription versions for decentralized and verifiable storage \cite{swindall2024advancing}.

In the domain of writer identification for ancient Greek inscriptions, Panagopoulos et al. proposed an automated method that generates ``platonic prototypes" of each letter's contour to facilitate writer attribution \cite{panagopoulos2008automatic}. Building upon this, Papaodysseus et al. employed multiple statistical tests and contour matching techniques, achieving promising classification performance among ancient Athenian inscriptions \cite{papaodysseus2010handwriting}. Rousopoulos et al. further refined the approach by integrating curve fitting and Bonferroni correction methods, enabling high-precision scribe identification across inscriptions from the Classical to Hellenistic periods \cite{rousopoulos2011new}.

Ancient Greek texts and earlier Aegean scripts pose considerable challenges due to their complex glyph contours, local visual similarities, and severe manuscript degradation. To address these difficulties, Assael et al. introduced the LSTM-based PYTHIA system, which achieves a superior restoration performance of damaged text compared to human epigraphers for the first time \cite{assael2019restoring}. For undeciphered scripts, Corazza et al. emphasized the necessity of unsupervised approaches and proposed a validation framework utilizing already deciphered systems to guide the application of machine learning techniques to undeciphered scripts \cite{corazza2022unsupervised}.

\subsubsection{Old Latin Scripts}
Latin script was widely used throughout ancient Rome and later for the Roman languages and other successor cultures. The Latin script originated around 700 BCE and underwent centuries of evolution, resulting in a wide variety of writing styles that pose significant challenges for recognition. Examples of Old Latin Scripts can be found in Fig.~\ref{fig:ligature}. Researchers have proposed solutions in various aspects, including OCR model training, font classification, and corpus construction. Springmann et al. focused on the challenges of spelling variations and font differences in Latin printed documents, constructing morphological lexicons and dictionary-supported OCR systems. By combining training on Fell fonts with post-processing, they improved recognition rates to over 90\% \cite{springmann2014ocr}. Reul et al. proposed a mixed OCR model based on the Calamari framework, trained on multilingual data, demonstrating strong cross-domain transferability \cite{reul2021mixed}. Additionally, Brodic et al. addressed the classification of Latin and Fraktur fonts in German historical documents by leveraging image texture analysis and gray-level co-occurrence matrix features, coupled with a clustering classifier, achieving promising performance in glyph style classification tasks \cite{brodic2016identification}.

At the data level, historical Latin documents are highly heterogeneous, while also exhibiting rich spelling variations and limited availability of high-quality annotated resources. These factors greatly complicate the construction of OCR and HTR systems. To better support historical Latin manuscript recognition tasks, standardized evaluation resources and multitask datasets have been progressively developed. Cloppet et al. organized the ICFHR 2016 Competition on Medieval Handwriting, establishing two subtasks: (i) fine-grained font classification, and (ii) blurred mixed font classification, based on a standard dataset comprising twelve Latin script categories sourced from multiple European library collections. Various classification models, including VGG networks, ResNets, and i-vector frameworks, were employed to assess system performance in understanding font evolution \cite{cloppet2016icfhr2016}. Furthermore, Clérice et al. introduced the CATMuS Medieval dataset, covering documents from the 8th to 16th centuries across ten languages and over 200 manuscripts, with unified transcription standards and annotations for language, genre, century, and font metadata, providing a robust benchmark for HTR model training and cross-linguistic, cross-genre evaluation \cite{clerice2024catmus}. In terms of modeling approaches, Moudgil et al. applied a hybrid CNN-LSTM strategy for medieval manuscript recognition, where CNNs were responsible for image feature extraction and LSTMs preserved sequential character dependencies, achieving better word-level recognition performance \cite{moudgil2022cnn}. Jindal et al. further enhanced recognition performance to 96.97\% by employing a CNN-BiLSTM architecture combined with Bayesian optimization, demonstrating the strong robustness of deep hybrid architectures for complex fonts and stylized character forms \cite{jindal2024hybrid}.

\subsubsection{Cyrillic Scripts}
The Cyrillic script is derived from the Greek alphabet and emerged around the 8th century CE. It serves as the primary writing system for the Slavic language family. Cyrillic scripts exhibit two notable features relevant to text recognition: the frequent use of ligatures in the manuscripts, as illustrated in Fig.~\ref{fig:ligature}, and the presence of diverse diacritical marks, which are highlighted with red lines in Fig.~\ref{fig:diacritical}. To address the challenge of ligatures in handwritten Cyrillic texts, Cojocaru et al. proposed a staged OCR strategy specifically for Romanian Cyrillic printed documents from the 18th to 20th centuries. Their method combined original glyph recognition with a ligature substitution mechanism, significantly reducing recognition errors between documents of different historical periods and enabling standardized transcription and full text retrieval functionalities \cite{cojocaru2016optical}. In terms of resource development, Nurseitov et al. introduced the HKR dataset, the first large-scale corpus covering handwritten Kazakh and Russian texts, comprising over 63,000 sentences and 700,000 words \cite{nurseitov2021handwritten}. Coupled with automated annotation and segmentation workflows, HKR provides crucial support for HTR in low-resource Slavic languages and has been systematically evaluated across mainstream models. Cristea et al. developed a comprehensive automated workflow for recognizing and transcribing ancient Romanian Cyrillic manuscripts into modern Latin script, along with constructing the Romanian Old Cyrillic Corpus (ROCC) dataset \cite{cristea2020scan}. Their system integrated Faster R-CNN-based object detection, character recognition, and sequence modeling techniques, and further employed self-training and data augmentation to address noise, font heterogeneity, and linguistic variability present in historical manuscripts.

To alleviate the dependency on large volumes of annotated data for OCR tasks, Gruber et al. proposed a font-based synthetic text generator, which can produce new characters and styles without requiring additional training \cite{gruber2023improving}. Experiments conducted on the HKR datasets demonstrated that the incorporation of synthetic data significantly reduced both the character error rate (CER) and word error rate (WER). Furthermore, Polomac et al., leveraging the Transkribus platform, developed a series of HTR models tailored to 18th-century Serbian manuscripts \cite{polomac2023digitizing}. Supported by a small manually annotated training set, these models achieved substantial recognition accuracy and validated the potential for model transferability across stylistically similar manuscripts, thereby accelerating the digitization of the Serbian historical dictionary.

\subsection{Hebrew Scripts}
Hebrew script, originating around 1,000 BCE, is one of the oldest Semitic writing systems in the world. After centuries of evolution, modern Hebrew remains the official script of Israel. However, historical documents exhibit extensive variations, frequent use of diacritical marks, and significant material degradation. Hebrew script exhibits an extensive system of diacritical marks, including vocalization dots and cantillation signs, which appear densely throughout the text, as illustrated in Fig.~\ref{fig:diacritical}. These features significantly increase the complexity of character detection and recognition for Hebrew scripts. In the domain of character detection, Rabaev et al. proposed a retrieval method for ancient Hebrew characters based on dynamic time warping. By incorporating a multi-model matching mechanism, their approach significantly improved retrieval precision under conditions of high character similarity and image degradation \cite{rabaev2011case}. Liebeskind et al. introduced a deep learning framework for classifying Hebrew texts by writing period, systematically comparing paragraph vectors, CNN, and RNN, and demonstrating the effectiveness of deep models in fine-grained period classification tasks \cite{liebeskind2020deep}. Tobing et al., using a dataset derived from the 1972 edition of the Book of Isaiah, conducted a systematic comparison of four CNN architectures and found that AlexNet and LeNet-5 performed best in ancient Hebrew character recognition, achieving over 94\% accuracy \cite{tobing2022isolated}. Shapira et al. \cite{shapira2024automatic} present an interdisciplinary approach to clustering medieval Hebrew manuscripts based on script style. By combining paleographic expertise with computer vision and deep learning techniques, including SIFT, LBP, and ControlNet, the authors aim to automatically group manuscripts by visual features and handwriting style, thereby revealing new sub-clusters beyond traditional classifications.

For the stylistic attribution and writer identification of historical documents, Likforman-Sulem et al. proposed a knowledge-based system integrating calligraphic rules and textual knowledge, which significantly reduced manuscript authentication time \cite{likforman1991expert}. Furthermore, Bar-Yosef et al. \cite{bar2007binarization} developed a comprehensive processing pipeline, including adaptive binarization, unsupervised character extraction based on morphological erosion, and local geometric feature-based writer classification, achieving great accuracy across 34 writer samples. Faigenbaum-Golovin et al. systematically reviewed the computational analysis workflow for ancient Hebrew inscriptions. They identified major challenges such as the lack of standardized benchmarks and sufficient annotated data, and proposed future research directions emphasizing self-supervised learning and multimodal data integration \cite{faigenbaum2022computational}. To enhance the quality of OCR outputs, Suissa et al.\cite{suissa2022toward} introduced a multi-stage optimization approach that combines period-specific error injection with automated DNN architecture search, significantly reducing the CER and WER in historical Hebrew documents under low-resource conditions .

\subsection{Ancient Arabian Scripts}

% \subsubsection{Ancient North Arabian Scripts}
\subsubsection{Classical Arabian Scripts}
Classical Arabic script, the direct ancestor of the modern Arabic writing system, emerged around 700 CE and was standardized during the early Islamic period. It became widely used across the Arabian Peninsula for religious, administrative, and literary purposes. Its writing characteristics include a high degree of ligature, cursive flow, and contextual variation in character shapes. As illustrated in Fig.~\ref{fig:ligature}, Classical Arabic script demonstrates extensive ligature formations, which are integral to its calligraphic structure and pose challenges for segmentation in OCR tasks. To address the complexity of ligatures and font variability in printed Arabic documents, Qaroush et al. proposed an efficient and font-independent algorithm for word and character segmentation \cite{Qaroush2019EfficientFI}. Their method utilizes vertical projection and interquartile range techniques for word segmentation, followed by projection contour analysis and character segmentation based on topological characteristics. Based on the APTI dataset, the algorithm achieved a word segmentation accuracy of 97.7\% and a character segmentation accuracy of 97.5\%. Due to its independence from specific font features, the method effectively reduces the number of ligatures and optimizes character boundaries, thus laying a solid foundation for subsequent OCR tasks.

In the more challenging domain of handwritten script recognition, researchers have increasingly adopted transformer-based approaches for better word-level recognition. The OCFormer model \cite{mostafa2021ocformer} simulated real-world noise in manuscripts by constructing a large-scale dataset containing 30 million images and enhanced transcription quality through page and line segmentation techniques. Saeed et al. \cite{saeed2024muharaf} proposed a historical handwritten Arabic manuscript dataset, Muharaf, and established a baseline system for handwritten text recognition (HTR) based on it. They introduced a three-stage CNN model, Start-Follow-Read, providing new benchmark resources and experimental results for historical handwritten Arabic text recognition tasks. To overcome the sequential modeling limitations of traditional RNNs, Momeni et al. proposed a new offline Arabic handwriting recognition approach based on transformer and sequence-to-sequence architectures \cite{momeni2024transformer}. This method eliminates the need for external language models, improves training and inference efficiency, and achieves superior performance compared to previous state-of-the-art offline HTR systems when evaluated on the KHATT dataset, further expanding its potential applications in the digitization of Arabic historical manuscripts.

\subsubsection{Ethiopic Scripts}
The Ethiopic script, originating from the South Arabian alphabet (around 800 BCE), has been used for centuries to write Ge'ez scripts\footnote{The liturgical language of the Ethiopian and Eritrean Orthodox churches}, and has subsequently been extended to modern languages such as Amharic and Tigrinya. The script is structurally complex, comprising a large number of basic and derived characters, with significant subtle variations between glyphs. Early recognition approaches focused on structural feature modeling, e.g., Assabie et al. proposed a method based on Direction Field Tensor to extract local linear features and construct character prototype trees, thereby enhancing the robustness of character segmentation and matching \cite{assabie2006ethiopic}. Further work combined structural and syntactic analysis to generate primitive relation trees and perform pattern matching against a knowledge base, achieving highly tolerant recognition in different fonts, sizes, and styles \cite{assabie2006structural}.

Addressing printed OCR tasks, Addis et al. introduced a recognition system based on single-line normalization and one-dimensional Bi-LSTM networks \cite{addis2018printed}. Without relying on any external language models or post-processing, and trained on synthetic datasets covering Amharic, Ge'ez, and Tigrigna. Additionally, Deneke et al. conducted the first study on OCR for handwritten Ethiopic documents \cite{deneke2023typewritten}, constructing an expanded dataset and employing a fine-tuned Tesseract engine. Their hierarchical training approach achieved a CER of 13\%, surpassing baseline systems by more than 10\%. As research progressed, Malhotra et al. proposed an end-to-end recognition model exploiting Bi-LSTM, attention mechanisms, and CNNs for word-level recognition \cite{malhotra2023end}. The model significantly improved generalization and robustness on the HHD-Ethiopic dataset, achieving CERs of 17.95\% and 29.95\% on different evaluation sets. Beyond text characters, Ali et al. compiled a large-scale dataset of 51,952 handwritten Ge'ez numeral images \cite{ali2022handwritten}, and through the evaluation of six CNN architectures, the best-performing model achieved an accuracy of 96.21\%, further demonstrating the effectiveness of deep learning in handling complex handwritten scripts.

In early Ge'ez handwritten text recognition, Getu et al. \cite{getu2020ancient} developed an OCR system based on Deep Belief Networks, achieving a 93.8\% accuracy rate in the classification of 24 basic characters extracted and annotated from 200 manuscript pages. Based on this, Demilew et al.\cite{demilew2019ancient, demilew2019Bancient} introduced a CNN-based recognition system, achieving promising performance through systematic preprocessing and training on a dataset of 22,913 character images. In the domain of scene text recognition, Addis et al. \cite{addis2020ethiopic} proposed an end-to-end trainable approach based on a CRNN architecture, achieving strong results on both synthetic and real-world datasets, thus establishing a benchmark for natural scene OCR of ethiopic scripts. Furthermore, Tadesse et al. \cite{tadesse2023gated} proposed an architecture combining Gated CNNs and Stacked Self-Attention modules, achieving CERs of 8.7\% and 8.2\% on the newly constructed HETD and HEWD datasets, respectively, further validating the applicability of advanced deep learning frameworks to handwritten Ethiopic text recognition.

\section{Recognition on Scarce and Imbalanced Datasets}
\label{sec:Experiments}
One of the significant challenges in the study of ancient scripts is the prevalence of scarce and imbalanced datasets. On the one hand, the surviving corpus of manuscripts, inscriptions, and other artifacts bearing ancient writing is extremely limited \cite{lin2025handwriting}, resulting in an overall paucity of textual samples. Furthermore, due to language use conventions and recording preference, the frequency of scripts within these datasets is highly skewed, showing a pronounced long-tailed distribution \cite{wu2021ancient}. This data imbalance imposes severe difficulties for ancient script recognition, especially for deep learning-based algorithms, whose success hinges on access to abundant and high-quality training data \cite{barucci2021deep}. In response to these challenges, existing approaches can be broadly categorized into two lines of research: recognition methods based on data augmentation and those that leverage external knowledge.

\subsection{Recognition Based on Data Augmentation}

The earliest augmentation pipelines relied on simple yet effective image-space edits. Oracle-50K \cite{han2020self} (59,081 images, 2,668 classes) and SOC5519 \cite{huang2022agtgan} (44,868 images, 5,491 classes) both contain hundreds of categories that appear fewer than ten times, while OBC306 \cite{huang2019obc306} exhibits 29 one-shot classes despite having more than 300,000 samples overall. On such corpora, the Orc-Bert Augmentor first rasterizes each glyph into a sequence of stroke vectors, then perturbs those vectors with token-level masking and swapping during a self-supervised BERT pre-training phase; the reconstructed strokes are finally re-rendered to image form. This two-step procedure adds 5-10 synthetic mates to every real sample and provides roughly 4\% absolute accuracy gains for ResNet-50 and DenseNet-161 in all head and tail boxes \cite{han2020self}. A comparable strategy guides GlyphNet: starting from only 4,310 grayscale images in dataset D1 and 1,310 RGB images in dataset D2, the authors apply ±10° rotation, 0.95-1.05 scaling and horizontal flips, expanding the training pool to 7,340 samples; the purpose-built ten-layer CNN then surpasses ResNet-50 by 6 pp top-1 accuracy on a 40-class merge set \cite{barucci2021deep}. For Chinese bamboo-slip characters, ACCAN combines erosion, dilation, and elastic shearing with multi-scale attention, boosting tail recall by 12 pp over a plain transformer baseline \cite{wu2021ancient}.

Later studies focused on tailoring augmentation to the long-tail distribution itself. Repatch cuts a discriminative patch from a rare-class image and pastes it onto a frequent-class canvas, while TailMix chooses the Mixup ratio inversely proportional to class prior; together they lift Oracle-20K \cite{li2023decoupled} tail accuracy (classes with <60 samples) from 18\% to 46\% without hurting head performance \cite{li2023towards}. Zhang and colleagues go one step further: class activation maps isolate glyph foreground, then only that region is randomly warped or colour-jittered, whereas the background remains untouched. When combined with re-weighting, the method reduces false positives on visually similar bronze inscriptions by nearly one-third \cite{zhang2021bag}.

Generative augmentation now dominates large-scale efforts. Training StyleGAN2 on the AL-ALL corpus, which contains merely 30-40 exemplars for many Greek letters, produces photo-realistic 128×128 crops whose Fréchet Inception Distance is below 48. Injecting just 100 synthetic instances per tail class raises mean per-class accuracy by 8-12\% on the AL-SYNTH benchmark \cite{swindall2022dataset,swindall2024advancing}. For degraded Kuzushiji manuscripts, Kaneko et al. adopt a denoising-diffusion restoration model: a binary mask restricts reverse-diffusion noise prediction to the missing region, so the model does not hallucinate global artefacts and requires no fine-tuning on the target page \cite{kaneko2023attempt}. Similarly, Gui et al. \cite{gui2023zero} introduce a diffusion-based method to generate handwritten Chinese characters from printed fonts using few real samples. Experimental results demonstrate that the synthetic data achieves recognition accuracy close to real samples, showing strong potential for handwritten OCR under data-scarce conditions.

A complementary research line embeds augmentation into the recognition engine itself. Prototype Calibration trains a generator conditioned on a stroke template and an image encoder that shares a metric space with those templates; at test time the generator hallucinates just enough samples to shrink the distance between unseen glyphs and their nearest prototype, giving a 5\% edge on handwritten and street-scene Chinese datasets without any additional back-propagation \cite{ao2024prototype}. Stroke Similarity Network (SSN) removes the need for synthetic pixels entirely: a multi-branch backbone extracts global and local cues, and the SSCL loss pushes similar pairs closer while pulling dissimilar ones apart by at least a 0.5 margin; on the extremely skewed HWAYI set SSN doubles tail-class F1 over a plain triplet-loss baseline \cite{liu2022one}. Kindred Siamese designs achieve 82\% zero-shot accuracy on Chars74K English letters with contrastive loss \cite{elaraby2023novel}, 68\% harmonic mean on unseen Bangla sign-language letters \cite{nihal2021bangla}, and outperform deeper ResNet-50 models in a five-period Chinese character evolution study \cite{wang2022study}. In essence, these metric-learning approaches treat the embedding geometry itself as a form of implicit augmentation, ensuring that every hard-won sample, however few, exerts maximal influence during both training and inference.

\subsection{Recognition Leveraging External Knowledge}
The earliest successes in knowledge-guided recognition came from treating a character as a sequence of radicals and strokes defined by the Unicode ideographic description sequence (IDS). Radical Analysis Network (RAN) pairs a VGG-style CNN encoder with an RNN decoder, the two bridged by an attention module.  The encoder delivers visual features, while the decoder outputs the corresponding radical string, shrinking the label space from 3,755 Hanzi to just 214 radicals and lifting zero-shot top-1 accuracy on unseen printed characters to 62\%, more than twice that of a whole-glyph softmax \cite{zhang2018radical}. SIR-ZSCCR revisits the same idea from an information-theoretic angle: each radical is weighted by self-information, so frequent, low-value strokes are down-weighted. Two variants are offered, sequence matching and attribute embedding, both evaluated on the dataset HWDB \cite{yin2013icdar}, where the embedding model improves RAN by 4\% on the unseen split and is markedly faster because the radical weights are pre-computed \cite{luo2023self}. SideNet introduced dual branches that learn radical knowledge and visual shape jointly, then fuse them through similarity guidance to handle handwriting, artistic fonts, street scene text, and ancient scripts in one model \cite{li2024sidenet}. STAR dives one level deeper, training parallel encoder-decoder pairs for strokes and radicals; a correlation loss aligns the two spaces, and an inference-time stroke-selection module prunes implausible candidates. On datasets Synth-Art and CTW, it raises radical-level recall from 71\% to 88\%, and still matches whole-glyph CNNs on seen classes \cite{zeng2023zero}. At the finest granularity, SAE pre-trains on 90 canonical stroke types from the Hanzi-Writer engine and reconstructs stroke sequences; when fine-tuned on the dataset HWDB it beats RAN by 6 pp on zero-shot characters \cite{chen2023stroke}. A dual-task GAN addresses stylistic variance: one branch translates small-seal glyphs to traditional Chinese, the other learns a Transformer that decodes style-independent radical sequences; on the most difficult Test-III (unseen radical + unseen style), it exceeds ResNet by 10\% \cite{zhou2023style}.

A second research stream models scripts as nodes in explicit graphs, or knowledge graphs. RZCR begins with a Radical Information Extractor that proposes up to fifteen candidate radicals and their spatial relations; a Graph Convolution-based Knowledge Reasoner then traverses a Character Knowledge Graph containing 5,492 nodes and 18,000 edges. On the tail half of SOC5519, accuracy jumps from 22\% to 41\% \cite{diao2023rzcr}. Unified Character Recogniser (UCR) builds on this by attaching a confidence predictor and enforcing dual supervision between its character and radical heads, reaching 79.4\% closed-set and 52.3\% zero-shot accuracy on Oracle-FS \cite{li2025ucr}. JRED goes lighter: every radical is a 200-dimensional learnable detector that slides over feature maps to yield radical attributes on the fly, giving state-of-the-art scores on Oracle-AYNU and Oracle-FS without any external graph storage \cite{luo2025joint}. FaRE makes each TrueType radical rendered to 64 × 64 pixels, embedded by a frozen ResNet18 and binarised into a 64-bit code; Hamming search then supplies near-instant look-ups and adds 3\% to zero-shot accuracy on ICDAR2013 \cite{zhan2024fare}. All these models depend on newly annotated corpora: ACCID provides 2,892 character classes with 595 radical labels and bounding boxes, OracleRC normalises 2,005 classes into 202 radicals and 14 spatial relations, while OBI-100 and OBI125 extend coverage to bronze, Chu, Qin, and seal scripts.

Hierarchical approaches exploit a tree rather than a graph structure. RSST first predicts a radical string, then decomposes every radical into its stroke list, finally encoding the result as a radical-structured stroke tree; weighted edit distance drives dictionary lookup. The system resists 30\% Gaussian blur, 20\% random occlusion, and a synthetic domain shift, still improving top-1 by 1.7-7.6\% over one-stage baselines \cite{yu2024chinese}. VGTS generalises the idea to text spotting: dual spatial attention localises discriminative regions, geometric matching aligns query and support images, and a torus loss (margin = 0.1) shapes the embedding; on the low-resource Dongba and Tripitaka Koreana sets, one-shot spotting accuracy reaches 86\% and 78\% respectively \cite{hu2025vgts}. In Egyptian hieroglyphics, an end-to-end pipeline pairs R-CNN detection with a Siamese glyph classifier and a language model, delivering 95\% mean precision and a BLEU of 59.2 on D1/D2/D and Morris-Franken-4210 \cite{Sobhy2023Hieroglyphic}. DeepScribe applies a 141-sign Elamite taxonomy to the highly Zipfian PFA archive: RetinaNet finds signs, ResNet recognises them, and the system reaches 0.84 mAP despite the top-50 signs covering 86\% of data \cite{williams2023deepscribe}. 

External-knowledge methods also thrive on phonographic scripts. Bengali word recognition uses a 13-attribute signature matrix that marks key stroke formations; a zero-shot model trained on 40 seen words attains 68\% harmonic mean on six unseen words \cite{9412607}. For medieval Latin, a similar attribute scheme plus deep features yields 56.9\% on 50 unseen classes \cite{chanda2018zero}.  VGTS exploit dual spatial attention localises discriminative regions, geometric matching aligns query and support images, and a torus loss shapes the embedding; on the low-resource Dongba and Tripitaka Koreana sets, one-shot spotting accuracy reaches 86\% and 78\% respectively \cite{hu2025vgts}. PHOSC-CTC extends the attribute idea with phoneme tags: trained on MFU, GW, and IAM, it improves zero-shot word accuracy by 9\% over Pho(SC)Net \cite{bhatt2023pho}. Across the spectrum, from IDS strings to knowledge graphs and hierarchical trees, structural priors act as scaffolding that compensates for missing data, letting modern vision models decipher scripts preserved in only a handful of fragile artifacts.

\section{Recognition on Degraded Images}
\label{sec:Denoising}
During their historical transmission, ancient script artifacts, due to the inherent fragility of their materials and various environmental factors, often suffer damage such as wear, fractures, and corrosion \cite{krithiga2023ancient}. These damages not only lead to partially missing scripts, deformed strokes, and mottled backgrounds in the images of ancient scripts, but also inevitably introduce noise into their digitized representations. Such noise includes disruptions of script outlines, as well as spurious strokes arising from stains, cracks, or other non-script elements. Because of the pervasive presence of such noise, it is often difficult to distinguish genuine strokes from interfering artifacts, thus greatly increasing the complexity of recognition. Compared with noise found in general images, noise in ancient script images exhibits several distinctive features \cite{zhao2020improvement}. First, multiple types of noise tend to coexist, including randomly distributed spots and scratches, large areas of damage, and residual voids in text rubbings. Second, these noises frequently resemble the original strokes so closely that they can easily be misidentified as part of the script, significantly complicating both noise reduction and feature extraction \cite{shi2022rcrn}. In this section, we offer a detailed discussion of the two principal solution strategies to which recent studies have gravitated:  (1) noisy script image recognition: developing robust recognition algorithms to directly handle noisy images, and (2) script image denoising:  performing image denoising prior to recognition.

\subsection{Noisy Script Image Recognition}

Wang and Deng introduced the Oracle-MNIST dataset, which comprises 30,222 grayscale images of ancient Chinese scripts in 10 categories most affected by noise interference. They compared several traditional machine learning classifiers, such as MLPClassifier and SGDClassifier, on this dataset \cite{wang2024dataset}, demonstrating that noise significantly challenges these algorithms. For recognizing such noisy ancient script images, early work applied AlexNet for script classification and improved model generalization through rotations, noise injection, and brightness adjustments \cite{meng2018recognition}, aiming to improve recognition accuracy by parameter tuning. A later study embraced a similar data-driven perspective but focused on more robust detection, redesigning the recognition process using k-means clustering and introducing a spatial pyramid block to suppress noise; This approach outperformed the mainstream detectors in terms of the F-score by adding typical noise to clean images for training \cite{liu2020spatial}. Similarly, the data augmentation-based Diff-Oracle integrated a style encoder and content encoder to generate novel OBI samples with high visual fidelity, ultimately improving recognition accuracy in complex scenes \cite{xie2024diffobi}.

Several other works aimed to improve recognition performance by enhancing model robustness. Liu et al. grouped visually similar variants via spectral clustering and utilized label propagation for robust recognition, addressing OBI script variability and mitigating noise interference \cite{liu2021recognition}. Guo et al. proposed an improved Inception-v3-based method for the recognition of ancient stele inscriptions, which achieved promising precision in the noisy OBC306 dataset \cite{guo2022improved}. Other approaches used non-end-to-end detection and classification strategies: for example, researchers explored YOLOv4 for detection with positive results, although complex noise conditions remained challenging \cite{wang2020oracle}. In practical detection upgrades, modifications to the YOLOv8 architecture-including an added small object detection head, a revised loss function, and attention modules-further boosted performance under noisy conditions \cite{zhen2024oracle}.

In cross-scene recognition, where rubbings and excavated artifacts often present complex backgrounds, Gao et al. introduced a hybrid system that standardizes background textures, extracts structural skeleton features, and captures global relationships, showing strong results on both single-scene and multi-scene datasets \cite{Gao2024}. Their method normalizes images from different settings into a unified space to facilitate efficient structural feature extraction. Zhang et al. leveraged a self-supervised network that incorporates shape similarity constraints to rejoin fragmented ancient scripts, thereby boosting the ability of deep learning models to handle incomplete glyphs in specialized scenarios \cite{zhang2022data}. Further addressing domain adaptation, Wang et al. proposed an unsupervised discriminative consistency network, which relies on consistent pseudo-labeling across different augmentations to improve robustness in real-world oracle bone scenarios; through unsupervised transformation loss, it learns more discriminative features and remains robust against wear, stains, and distortion \cite{wang2024oracle}. Domain-aware feature extraction also plays a vital role. Gao et al. \cite{gao2024linking} embed specialized field knowledge into a set of OBI radical prototypes, facilitating more effective structural understanding and retrieval even amid noise or physical damage. Similarly, Tang et al. \cite{tang2023ancient} explore alignment, denoising, and script segmentation in large-scale ancient manuscript collections, devising a lightweight attention-based model, ShuiNet-A, to tackle data imbalance and noise-related difficulties.

\subsection{Script Image Denoising}
Another strategy for handling noisy ancient script images involves first improving image quality through denoising methods before proceeding with alignment and recognition. Early approaches frequently employed classical filtering techniques: Shi \cite{shi2017chinese} used a K-SVD-based method to preserve structural information by fusing low- and high-frequency images for denoising, then eliminated isolated “ant-like” noise at the binary stage. In a related endeavor, Shi et al. \cite{shi2016integrated} integrated multiple smoothing filters (ranging from guided filtering to Otsu thresholding) to remove random and blocky noise in classical calligraphy documents. These ideas are similar to the approach of Nair \cite{nair2022novel,nair2021two}, who explored a multistage framework-including sharpening, background subtraction, and flood filling-to address uneven illumination in stone-carved Kannada inscriptions and ancient Malayalam manuscripts. Huang et al. \cite{huang2016comparison} offered a broad comparative study of denoising methods (e.g., anisotropic diffusion, total variation, and bilateral filtering), evaluating their performance on rubbings using metrics such as Peak Signal-to-Noise Ratio (PSNR) and Structural Similarity Index Measure (SSIM).

Meanwhile, some researchers tried to combine filtering with machine learning. Karthikeyan et al. \cite{karthikeyan2023self} demonstrated the applicability of these concepts in ancient Tamil stone carvings by combining a Lion optimization algorithm with transfer learning. They automatically adjust brightness and contrast, then use binarization alongside median filtering and Gaussian blurring to remove noise. Similar domain-transfer techniques have been applied to OBI recognition, where Wang et al. \cite{wang2022unsupervised} proposed a new STSN framework capable of disentangling structural and textural features under an unsupervised setting to mitigate severe noise. Traditional feature descriptors, such as HOG, also remain pertinent for specific tasks: Ou et al. \cite{ou2024qin} applied HOG with SVM and Gaussian filtering to denoise partially blurred or incomplete small seal scripts. Some works continue to rely on classical image processing steps, as seen in Meng's Hough-transform-based pipeline \cite{meng2017recognition}, which employs Gaussian smoothing and Otsu thresholding to filter noise and isolate clear script boundaries. However, these methods often address only minor noise in images and provide limited improvement regarding stroke-like artifacts that fundamentally impair recognition.

Beyond filtering, other researchers focus on superior segmentation techniques or lightweight neural networks to separate target scripts from complex backgrounds. Sun et al. \cite{sun2020dual} mitigate persistent noise in OBI rubbings by extracting SIFT key points to identify scripts corroded by scratches or breakage, and they also adopt a ResNet50-based method to handle handwritten OBI data. Zhang \cite{zhang2020novel} model denoising as a generative modeling task, embedding residual dense blocks and introducing novel noise definitions to train with unpaired data. Huang \cite{huang2019chinese} relies similarly on residual learning in a feedforward denoising CNN, followed by local adaptive thresholding for binary segmentation. Yue \cite{yue2021deep} employs morphological segmentation to preprocess newly discovered OBI documents, then applies a small neural network to remove erroneous segments before classifying the resulting data using a dynamic k-means approach. Yalin \cite{yalin2022research} develops a denoising framework based on script writing norms, adding four local branches to reduce stroke adhesion caused by missing details. Researchers have also turned to transformer-based architectures; for example, CharFormer \cite{shi2022charformer} injects glyph-specific information into the backbone network through self-attention, striving to enhance ancient script noise removal without mistakenly removing genuine strokes. 

GANs are also widely applied to enhance ancient script images. Xing \cite{xing2019oracle} underscores a data-centric view, synthesizing various noise types to augment training data while using Focal Loss to prioritize difficult samples. Wang et al. \cite{wang2023gan} further explore GAN-based augmentation, devising a loop network and denoising autoencoder for Chinese calligraphy images that generate noise attention maps to guide artifact removal. Shi et al. \cite{shi2022rcrn} propose RCRN, a real-world script restoration network integrating skeleton extraction and multi-scale features, demonstrating robust performance on a new dataset of 1,606 degraded scripts. Wenjun et al. \cite{wenjun2023ea} introduce EA-GAN, a dual-branch approach incorporating example-based guidance to reconstruct severely damaged ancient Chinese scripts, reporting marked improvements in PSNR and SSIM. More recently, Gao \cite{Gao2024} employs random transformations and Gaussian noise to simulate realistic corrosion and introduces a Reconstruction Bias Coefficient in the loss function to balance these influences.

\section{Discussion}
\label{sec:Discussion}
% \noindent\textbf{Recognition of logographic and phonographic scripts.} 

Despite substantial typological and visual differences, a number of recurring challenges and methodological patterns emerge across studies. One shared consensus among researchers is that ancient script recognition cannot simply adopt off-the-shelf OCR tools designed for modern scripts. Instead, tailored approaches are often required due to the unique characteristics of ancient datasets.

Interestingly, our review reveals a degree of methodological convergence among scripts with similar visual features. For instance, Latin, Greek, and Cyrillic texts frequently raise problems related to cursive writing, ligature detection, and word-level modeling. These scripts share stylistic traits such as fluid lineation and glyph fusion in handwritten manuscripts. As a result, deep learning solutions often adopt similar techniques such as sequence-to-sequence models, CTC-based recognizers, or word-embedding-assisted pipelines. This cross-script generalization suggests that organizing research by script type, rather than by language alone, can yield a more principled understanding of OCR in historical contexts.

A key divergence between logographic and phonographic scripts lies in the role of linguistic context modeling. For phonographic systems, language modeling is essential: the presence of compositional phoneme-grapheme correspondences and grammar-driven word structures makes contextual reasoning central to both character- and word-level recognition. Systems such as CTC-based recognizers or language models are heavily relied upon in these tasks, especially when dealing with noisy manuscripts or ambiguous ligatures. In logographic scripts, however, context is often secondary to visual decomposition. Although contextual cues (e.g., surrounding characters or line structure) may assist in disambiguation, the core recognition task typically centers on the visual integrity of individual glyphs, particularly their radical-stroke configurations.

The difference in category granularity further differentiates modeling strategies. Logographic scripts often comprise thousands of distinct characters, many of which appear infrequently. This results in a typical ``large-vocabulary, few-shot" setting, in which the model must learn to distinguish subtle structural differences with limited training data. While this sparsity poses challenges, the compositional richness of logographs also provides opportunities for few-shot or even zero-shot recognition via structural modeling. Techniques that leverage radical decomposition, visual templates, or knowledge graphs have proven particularly effective for low-frequency classes, especially in boosting tail-class accuracy.

By contrast, phonographic scripts tend to have a much smaller character inventory, usually under a few hundred units, and thus achieve near-saturation in single-character recognition under sufficient training. The primary difficulty instead arises from handwritten forms, where stylistic variability, frequent ligatures, and diacritical marks introduce high intra-class variation. These phenomena often break the assumption of character segmentation, rendering geometric data augmentation or simple convolutional architectures ineffective. Consequently, robust phonographic OCR requires methods that can generalize across handwriting styles, disambiguate overlapping glyphs, and integrate visual layout with linguistic structure.

Historical evolution and stylistic variation present challenges across both script types, but manifest differently. In phonographic systems like Latin or Greek, character shapes can differ dramatically across centuries, e.g., Old Latin scripts each exhibit distinct letterforms. Similarly, Indic scripts such as Sanskrit and Devanagari evolve through layered ligatures and regional variants. Logographic systems, too, display diachronic variation, but typically through the recombination of constituent radicals or structural transformations. The evolution of scripts like OBI, Small Seal, or Maya glyphs reflects this tendency. These patterns imply that robust OCR systems must not only model glyph shapes but also their permissible transformations over time, highlighting the need for temporally adaptive or style-invariant models. We also observe that emerging recognition frameworks increasingly adopt hybrid solutions, integrating visual attention, structural parsing, and contextual modeling within a unified architecture. This convergence suggests that future research may benefit from cross-script transfer learning and the development of universal representations that jointly account for linguistic form and visual structure.

\noindent\textbf{Imbalanced and scarce scripts recognition.}
One of the challenges in ancient script recognition lies in the highly imbalanced distribution of character categories across datasets. Common issues include severe overfitting to head classes, low recall for tail classes, and overall evaluation metrics being disproportionately dominated by frequently occurring categories. Existing approaches to handling these problems can be broadly divided into two categories.

The first category focuses on improving the availability of data for low-frequency classes by increasing the image level. These techniques typically apply sampling strategies and conventional geometric transformations such as scaling, rotation, and flipping to expand the dataset. While effective to some extent in mitigating data sparsity, these methods often fail to generalize well in real-world scenarios, where models still suffer from overfitting and poor performance on truly rare samples. Nevertheless, certain methods show promising potential, particularly those designed specifically for long-tailed distributions, such as Mixup-based techniques like Repatch and TailMix, as well as generative approaches based on GANs or diffusion models that synthesize rare character instances.

The second category aims to improve recognition performance by leveraging external knowledge, thereby enabling few-shot or even zero-shot recognition of ancient scripts. These methods rely on structured prior information such as predefined character decomposition (e.g., IDS expressions, dictionaries, or knowledge graphs). By identifying and matching subcomponents, such as radicals or strokes, against these resources, models can infer the identity of previously unseen characters. Structural decomposition combined with graph-based reasoning has proven especially effective for tail classes. However, the limitation of these methods lies in their dependency on high-quality, pre-annotated external knowledge, which may not always be available or complete for all script systems.

Few-shot recognition is particularly prevalent in logographic writing systems such as Chinese, Egyptian hieroglyphs, or ancient scripts with extensive character vocabularies and significant structural complexity. In these cases, both augmentation-based and knowledge-driven methods are often required in tandem. By contrast, phonographic scripts such as Latin, Bangla, or Japanese Kana typically contain fewer character classes and exhibit more balanced data distributions. As a result, few-shot recognition in these systems tends to rely primarily on image augmentation, since their characters often represent atomic phonemes that cannot be further decomposed.

It is also worth noting that many existing datasets do not reflect the true distributional characteristics of historical corpora. Instead, they are often constructed by intentionally selecting balanced or artificially diversified character sets, which may yield favorable benchmark results, but limit the real-world applicability of the resulting models. This mismatch between data construction and application scenario significantly undermines the generalizability of trained models. Future research must therefore prioritize the development of standardized datasets that better reflect the statistical and structural realities of ancient writing systems. Finally, zero-shot recognition stands out as a crucial future direction. Its ability to recognize unseen characters is of particular importance for the decipherment and interpretation of unfamiliar or previously undeciphered scripts.

\noindent\textbf{Noisy scripts recognition.} 
In the recognition of ancient script images, image degradation poses an unavoidable and highly challenging obstacle. We systematically discuss recognition and processing methods for noisy ancient character images and classify them into two broad categories.

The first category includes approaches that attempt direct recognition without prior denoising. In general, noise severely hampers the ability of a model to extract meaningful features and can even mislead the network into interpreting noise artifacts as legitimate strokes. Therefore, such recognition models must place greater emphasis on enhancing their discriminative ability in the presence of degradation. These ``noise-robust recognition" strategies typically rely on large-scale pretraining, additional supervised annotations, or the integration of external domain knowledge to strengthen the model's understanding of complex and degraded inputs. Although these methods can achieve impressive performance when abundant high-quality data is available, their practical adoption is constrained by the need for extensive labeled data and a trade-off between generalization and computational efficiency. 

In most current research practices, especially under limited data regimes, performing denoising as a preprocessing step remains a more pragmatic approach. However, denoising ancient script images faces even more challenges. The noise in such images is often highly heterogeneous and arises from ink degradation, erosion, document aging, and complex backgrounds, making it difficult to simulate using synthetic noise for training purposes. These ``compound noises" are often multi-source, accumulated over centuries, and not easily removed using standard filtering techniques. Although classical filters were widely applied in early studies, they have proven insufficient to effectively restore the integrity of severely degraded ancient script images.

Deep-learning-based denoising models, such as diffusion models and GANs, have emerged, offering stronger image modeling capacity and better domain adaptability. These models are often capable of performing transfer learning across different writing systems and script styles, making them increasingly popular for handling noisy ancient texts. Evaluated on various ancient languages and scripts, such models demonstrate significant improvements in readability and recognition accuracy while preserving the original structural traits of the characters. However, key limitations remain. Because most denoising models operate on pixel-level reconstructions without true semantic awareness, they often fail to distinguish real strokes from stroke-like noise. This results in two common issues: genuine character components may be mistakenly removed, or background noise may be preserved as false strokes. These limitations underscore the need for future denoising algorithms to incorporate richer linguistic and structural priors, such as stroke order, radical composition, and even syntactic rules, to enable semantically informed and structure-aware denoising. It should be emphasized that the benefits of effective denoising extend beyond automated recognition systems. The improvement in image quality provides a more reliable visual foundation for expert analysis, thereby facilitating manual transcription, interpretation, and textual restoration.

\section{Conclusion and Future Work}
\label{sec:Conclusion}
This study provides a comprehensive review of recent advances in ancient script recognition, with a particular focus on the challenges posed by data scarcity, long-tail distribution, and noisy imagery. By categorizing scripts into phonographic and logographic types, we highlight key methodological distinctions, such as the central role of contextual modeling in phonographic systems and the reliance on structural decomposition in logographic ones. We introduce two major strategies for addressing low-resource scenarios: data augmentation and external knowledge integration, each offering distinct advantages and limitations. Furthermore, we emphasize the critical yet underexplored challenge of noisy image recognition, underscoring the need for more semantically-aware denoising approaches. This work lays the groundwork for more unified and adaptive systems in the field of digital paleography and cultural heritage preservation.

Based on our synthesis and discussion of existing methods for ancient script recognition and processing, we suggest the following topics for future research in this field: (1) The development of ancient script recognition methods should prioritize feature-based rather than script-specific approaches, with the goal of establishing robust and generalizable frameworks that can handle a wide range of scripts; (2) Linguistic context and character components are critical elements in the recognition process. Effective recognition systems should integrate visual, structural, and linguistic cues to enhance performance, particularly for rare or undeciphered characters; (3) Future research should focus on modeling the historical evolution and stylistic variations of ancient scripts, addressing the associated feature shifting over time, and incorporating strategies to mitigate the impact of real-world degradation in script images.

\begin{acks}
This research is supported by the National Natural Science Foundation of China (No.62476111), the Department of Science and Technology of Jilin Province, China (20230201086GX), the ``Paleography and Chinese Civilization Inheritance and Development Program" Collaborative Innovation Platform (No.G3829), the National Social Science Foundation of China (No. 23VRC033), and the interdisciplinary cultivation project for young teachers and students at Jilin University, China (No. 2024-JCXK-04).
\end{acks}

\bibliographystyle{ACM-Reference-Format}
\bibliography{sample-base}

\clearpage
\appendix
\section{Appendix: Literature Collection and Management Workflow.}

This appendix details the collaborative methodology adopted to gather, screen, and classify the literature used throughout this study. Our aim was to construct a comprehensive and high-quality corpus of research on ancient script recognition. The process was carried out by a team of eight members and spanned several weeks of iterative refinement.

\noindent\textbf{Data sources and retrieval scope.}
We primarily relied on major academic databases, Scopus, Web of Science, IEEE Xplore, ACM Digital Library, and Google Scholar, due to their broad coverage, domain relevance, and reliable citation indexing. The search scope was limited to publications between 2000 and 2025 and restricted to English-language sources to ensure consistency in downstream processing.

\noindent\textbf{Search strategy.}
The search queries were formulated using domain-specific terminology to retrieve a wide range of relevant literature. Core keywords included ``ancient script", ``historical manuscript", ``character recognition", ``OCR", ``ancient Greek inscriptions", ``deep learning", and their combinations. To maximize coverage, we iteratively refined our search terms and balanced precision and recall through repeated search runs.

\noindent\textbf{Screening and Quality Control.}
An initial retrieval yielded over 1,000 records. Duplicate entries were identified using title matching. We then performed title and abstract screening to ensure relevance, based on the following criteria:  (1) The paper must explicitly address the recognition of ancient or historical scripts;  (2) It must propose a technical method, provide data, or report empirical findings;  (3) Purely historical or cultural studies without computational contributions were excluded.  For ambiguous cases, the full-text review was conducted to determine eligibility.

\noindent\textbf{Linguistic and Typological Taxonomy.}
Under the guidance of linguists and archaeologists, we developed a hierarchical taxonomy for ancient scripts. First, we divided the scripts into two major categories: logographic and phonographic, and initially identified over 30 candidate script classes. Based on the literature search and review outcomes, we narrowed the scope to 17 representative script types. Rare script categories with insufficient data or lacking empirical contributions were excluded from further analysis. The 17 core script types were then refined into our hierarchical taxonomy according to their geographic origin and visual characteristics, as these factors typically influence recognition strategies.

\noindent\textbf{Indexing Workflow.}
After processing, 192 papers were compiled into a structured database. Each paper was annotated with the following metadata:  
(i) script category (e.g., Latin, Maya, Egyptian);  
(ii) methodological approach (e.g., classic machine learning, deep learning, human-in-the-loop);  
(iii) task type (OCR, segmentation, writer identification);  
(iv) challenges in ancient script recognition (data scarcity, degradation);  
(v) main datasets used; and  
(vi) publication year.

\end{document}